\documentclass{article} % For LaTeX2e
\usepackage[preprint]{tmlr}
\usepackage{amsmath,amsfonts,bm}

\def\eqref#1{equation~\ref{#1}}
\def\1{\bm{1}}

\DeclareMathAlphabet{\mathsfit}{\encodingdefault}{\sfdefault}{m}{sl}
\SetMathAlphabet{\mathsfit}{bold}{\encodingdefault}{\sfdefault}{bx}{n}

\usepackage{xcolor}
\definecolor{citec}{HTML}{882810}
\definecolor{refc}{HTML}{4658cf}
\definecolor{enp}{HTML}{50ace9}
\definecolor{urlc}{HTML}{39a85c}
\definecolor{tablebg}{HTML}{E6E6E6}
\usepackage[colorlinks,
            linkcolor=refc,
            anchorcolor=blue,
            urlcolor=urlc,
            citecolor=citec]{hyperref}
\usepackage{url}
\usepackage{wasysym}
\usepackage{multicol}
\usepackage{latexsym}
\usepackage{booktabs}
\usepackage{amssymb}
\usepackage{amsmath}
\usepackage{subfig}
\usepackage{graphicx}
\usepackage{multirow}
\usepackage[ruled,linesnumbered]{algorithm2e}
\usepackage{colortbl}
\usepackage{rotating}
\usepackage[inkscapelatex=false]{svg}
\usepackage{diagbox}
\usepackage{wrapfig}
\usepackage{colortbl}
\usepackage{nicematrix}
\usepackage{makecell}
\usepackage{bbm}

\title{StaICC: Standardized Evaluation for Classification Task in In-context Learning}

\author{
\name \hspace{-1.5mm}Hakaze Cho
\addr $\vert$ Japan Advanced Institute of Science and Technology \\
\name Naoya Inoue 
\addr $\vert$ Japan Advanced Institute of Science and Technology, RIKEN \\ \name
\addr {\textbf{Correspondence to:} \texttt{yfzhao@jaist.ac.jp}.}
}

\newcommand{\M}{\textsc{StaICC}}
\newcommand{\MN}{\textsc{StaICC-Normal}}

\newcommand{\MD}{\textsc{StaICC-Diag}}

\begin{document}

\maketitle

\begin{abstract}
Classification tasks are widely investigated in the \textbf{I}n-\textbf{C}ontext \textbf{L}earning (ICL) paradigm. However, current efforts are evaluated on disjoint benchmarks and settings, while their performances are significantly influenced by some trivial variables, such as prompt templates, data sampling, instructions, etc., which leads to significant inconsistencies in the results reported across various literature, preventing fair comparison or meta-analysis across different papers. Therefore, this paper proposes a standardized and easy-to-use evaluation toolkit (\M) for in-context classification. Including, for the normal classification task, we provide \MN, selecting 10 widely used datasets, and generating prompts with a fixed form, to mitigate the variance among the experiment implementations. To enrich the usage of our benchmark, we also provide a sub-benchmark \MD~for diagnosing ICL from several aspects, aiming for a more robust inference processing.
\end{abstract}
\begin{center}
    \vspace{-0.5\baselineskip}
    \textbf{Library released:} \url{https://github.com/hc495/StaICC}
\end{center}

\section{Introduction}

\textbf{I}n-\textbf{C}ontext \textbf{L}earning (ICL)~\citep{dong2022survey} is an emerging forward-calculation-only few-shot paradigm using \textbf{L}anguage \textbf{M}odels (LMs) without parameter updating. ICL paradigm can be used on various task forms~\citep{min2022metaicl, min2022rethinking}, where \textit{classification}s with finite and fixed label space are concise and critical tasks for calibrating LMs for ICL~\citep{zhao2021calibrate, han2023prototypical, jiang2023generative, fei2023mitigating, zhou2024batch, cho2024token}, aligning LMs towards ICL objective~\citep{wei2022finetuned, min2022metaicl, wei2023symbol, iyer2022opt}, and investigating the inner principle of ICL~\citep{min2022rethinking, yoo2022ground, dai2023can, han2023understanding, wang2023label, kossen2024context, cho2024revisiting}.

However, previous works use disjoint benchmarks to evaluate the ICL performance on classification tasks (Table~\ref{tab:dataset_usage}), and performance even on the same model and dataset varies (Fig.~\ref{fig:result_disjoint}) due to non-essential factors, such as the prompt template, data sampling, demonstration order, etc. Due to such an inconsistency, in current ICL research, scholars usually need to repeat the baseline experiments in a consistent setting to compare the results with previous works, even if these experiments are quite costly. This greatly increases the difficulty of comparing or meta-analyzing the results in the literature, causing a limited understanding of how ICL performance varies across models and additional algorithms for improving ICL.

To this end, in this paper, we propose \M, a standardized and easy-to-use toolkit for evaluating the classification performance of ICL. Concretely, we select 10 widely used datasets on single-sentence classification tasks to unify the usage of datasets and fix the prompt template, data sampling, and demonstration orders to generate stable-over-trial test inputs, to eliminate the inconsistency caused by these experimental settings. Moreover, to enrich the usage of \M, we propose~\MD, for diagnosis data consisting of prediction bias, prompt sensitivity, label-noise robustness, etc., to help for a closer look and improvement of ICL inference methods.

% Moreover, to explore ICL abilities on prospective tasks, we propose~\MH, consisting of 4 subtasks of (1) Multi-token label classification, (2) Long text classification, (3) Multi-input classification, (4) More-way classification. and~\MD~for diagnosis data consisting of prediction bias, prompt sensitivity, etc., to help the improving of ICL methods.

Based on \M, we extensively measure the ICL capability on 29 modern LMs, and observe clear scaling laws of ICL classification performance against the model parameter numbers, confirming the robustness of \M~and responding to our initial motivation of a consistent inference evaluation. Moreover, we fairly evaluate the performance of 10 inference methods on \M, providing baseline and reference data for subsequent research.

% Please add the following required packages to your document preamble:
% \usepackage{booktabs}
\begin{table}[t]
\setlength{\tabcolsep}{5pt}
\renewcommand{\arraystretch}{1.0}
\setlength{\abovecaptionskip}{2pt}
\centering
\caption{\small Usage of datasets and metrics in 29 papers related to ICL. Significant variances are found in datasets and metrics usage in the literature. \textbf{Notation}: (Blank): un-used; \LEFTcircle: measured by Accuracy; \RIGHTcircle: measured by Macro-F1; \CIRCLE: measured by both Accuracy and F1; \Circle: measured by other metrics.}
\label{tab:dataset_usage}
\resizebox{1\columnwidth}{!}{
\begin{NiceTabular}{@{}ccccccccccccccc@{}}
\toprule
\rowcolor{tablebg}
 \textbf{Literature}&
  \textbf{SST$_2$} &
  \textbf{RTE} &
  \textbf{AGN} &
  \textbf{TREC} &
  \textbf{Subj.} &
  \textbf{MR} &
  \textbf{DBP} &
  \textbf{SST$_5$} &
  \textbf{MRPC} &
  \textbf{CB} &
  \textbf{FinP} &
  \textbf{WNLI} &
  \textbf{TEE} &
  \textbf{\textit{Others}} \\ \midrule
\cite{kossen2024context} &
  \LEFTcircle &
  \LEFTcircle &
  \LEFTcircle &
   &
  \LEFTcircle &
   &
   &
   &
  \LEFTcircle &
   &
  \LEFTcircle &
  \LEFTcircle &
   &
   +2\\
\cite{wang2023label} &
  \Circle &
   &
  \Circle &
  \Circle &
   &
   &
   &
   &
   &
   &
   &
   &
   &
   +1\\
\cite{gu2023pre} &
  \LEFTcircle &
  \LEFTcircle &
  \LEFTcircle &
   &
  \LEFTcircle &
  \LEFTcircle &
   &
  \LEFTcircle &
   &
  \LEFTcircle &
   &
   &
   &
   \\
\cite{han2023understanding} &
  \LEFTcircle &
   &
  \LEFTcircle &
   &
   &
   &
   &
   &
   &
   &
   &
   &
   &
   +4\\
\cite{pan2023context} &
  \LEFTcircle &
   &
   &
   &
   &
   &
   &
   &
  \LEFTcircle &
   &
  \LEFTcircle &
   &
  \LEFTcircle &
  +9 \\
\cite{yoo2022ground} &
  \CIRCLE &
  \CIRCLE &
   &
  \CIRCLE &
   &
   &
   &
   &
  \CIRCLE &
   &
  \CIRCLE &
  \CIRCLE &
  \CIRCLE &
  +8 \\
\cite{min2022rethinking} &
  \CIRCLE &
  \CIRCLE &
   &
  \CIRCLE &
   &
   &
   &
   &
  \CIRCLE &
   &
  \CIRCLE &
  \CIRCLE &
  \CIRCLE &
  +7 \\
\cite{jiang2023generative} &
  \RIGHTcircle &
  \RIGHTcircle &
  \RIGHTcircle &
  \RIGHTcircle &
  \RIGHTcircle &
  \RIGHTcircle &
  \RIGHTcircle &
  \RIGHTcircle &
   &
  \RIGHTcircle &
   &
   &
   &
   +3 \\
\cite{zhou2024batch} &
  \LEFTcircle &
  \LEFTcircle &
   &
   &
   &
   &
   &
   &
  \LEFTcircle &
  \LEFTcircle &
   &
   &
   &
   +7 \\
\cite{han2023prototypical} &
  \LEFTcircle &
  \LEFTcircle &
  \LEFTcircle &
  \LEFTcircle &
  \LEFTcircle &
  \LEFTcircle &
  \LEFTcircle &
  \LEFTcircle &
   &
   &
   &
   &
   &
   +1\\
\cite{fei2023mitigating} &
  \RIGHTcircle &
  \RIGHTcircle &
  \RIGHTcircle &
  \RIGHTcircle &
  \RIGHTcircle &
  \RIGHTcircle &
  \RIGHTcircle &
  \RIGHTcircle &
   &
  \RIGHTcircle &
  \RIGHTcircle &
  \RIGHTcircle &
  \RIGHTcircle &
   +9\\
\cite{zhao2021calibrate} &
  \LEFTcircle &
  \LEFTcircle &
  \LEFTcircle &
  \LEFTcircle &
   &
   &
  \LEFTcircle &
   &
   &
  \LEFTcircle &
   &
   &
   &
   +3\\
\cite{zhao2024noisyicl} &
  \CIRCLE &
  \CIRCLE &
   &
   &
   &
   &
   &
   &
  \CIRCLE &
   &
  \CIRCLE &
   &
  \CIRCLE &
   +6\\
\cite{min2022metaicl} &
  \CIRCLE &
  \CIRCLE &
   &
  \CIRCLE &
   &
  \CIRCLE &
   &
   &
  \CIRCLE &
   &
   &
  \CIRCLE &
   &
  +37 \\
\cite{wei2023symbol} &
   &
   &
   &
   &
  \LEFTcircle &
   &
   &
   &
   &
   &
   &
   &
  \LEFTcircle &
  +9 \\
\cite{wei2023larger} &
  \LEFTcircle &
  \LEFTcircle &
   &
   &
  \LEFTcircle &
   &
   &
   &
   &
   &
  \LEFTcircle &
  \LEFTcircle &
   &
+4\\
\cite{shi2024larger} &
  \LEFTcircle &
  \LEFTcircle &
   &
   &
  \LEFTcircle &
   &
   &
   &
   &
   &
   &
  \LEFTcircle &
   &
   +1\\
\cite{chen2023relation} &
   &
   &
  \Circle &
   &
   &
  \Circle &
   &
   &
   &
   &
   &
   &
   &
  +8 \\
\cite{min2022noisy} &
  \LEFTcircle &
   &
  \LEFTcircle &
  \LEFTcircle &
  \LEFTcircle &
  \LEFTcircle &
  \LEFTcircle &
  \LEFTcircle &
   &
   &
   &
   &
   &
  +4 \\
\cite{li2023finding} &
  \LEFTcircle &
   &
  \LEFTcircle &
  \LEFTcircle &
  \LEFTcircle &
  \LEFTcircle &
  \LEFTcircle &
  \LEFTcircle &
   &
   &
   &
   &
   &
  +1 \\
\cite{wei2022finetuned} &
  \LEFTcircle &
  \LEFTcircle &
   &
   &
   &
   &
   &
   &
  \LEFTcircle &
  \LEFTcircle &
   &
  \LEFTcircle &
   &
  +27 \\
\cite{cho2024token} &
   &
  \CIRCLE &
  \CIRCLE &
   &
   &
  \CIRCLE &
   &
   &
   &
   &
  \CIRCLE &
   &
  \CIRCLE &
  +4 \\
\cite{lu2022fantastically} &
  \LEFTcircle &
  \LEFTcircle &
  \LEFTcircle &
  \LEFTcircle &
  \LEFTcircle &
  \LEFTcircle &
  \LEFTcircle &
  \LEFTcircle &
   &
  \LEFTcircle &
   &
   &
   &
  +2 \\
\cite{xu2023knn} &
  \LEFTcircle &
  \LEFTcircle &
  \LEFTcircle &
  \LEFTcircle &
  \LEFTcircle &
  \LEFTcircle &
  \LEFTcircle &
   &
   &
  \LEFTcircle &
   &
   &
   &
  +2 \\
\cite{abbas2024enhancing} &
  \LEFTcircle &
  \LEFTcircle &
  \LEFTcircle &
  \LEFTcircle &
  \LEFTcircle &
   &
  \LEFTcircle &
  \LEFTcircle &
   &
   &
   &
   &
   &
  \\
\cite{su2023selective} &
   &
  \LEFTcircle &
   &
   &
   &
   &
  \LEFTcircle &
  \LEFTcircle &
  \LEFTcircle &
   &
   &
   &
   &
  +1 \\
\cite{shi2024incontext} &
  \LEFTcircle &
   &
  \LEFTcircle &
   &
   &
   &
  \LEFTcircle &
   &
   &
   &
   &
   &
   &
  +4 \\
\cite{hao2022structured} &
  \LEFTcircle &
  \LEFTcircle &
  \LEFTcircle &
  \LEFTcircle &
  \LEFTcircle &
  \LEFTcircle &
  \LEFTcircle &
  \LEFTcircle &
   &
  \LEFTcircle &
   &
   &
   &
   \\
\cite{li2023unified} &
  \LEFTcircle &
   &
  \LEFTcircle &
  \LEFTcircle &
   &
  \LEFTcircle &
  \LEFTcircle &
  \LEFTcircle &
   &
   &
   &
   &
   &
  +4 \\ \midrule
count: 29 &
  25 &
  20 &
  18 &
  15 &
  14 &
  13 &
  13 &
  11 &
  9 &
  9 &
  8 &
  8 &
  7 &
  - \\ \bottomrule
\end{NiceTabular}}
\vspace*{-1.4\baselineskip}
\end{table}

% Moreover, we confirm the discriminability of \M~by calibrating it on scaling laws.

% Then, from each dataset, we create stable and homogenous \textit{demonstration}, \textit{calibration}, and \textit{test} splits, and assemble prompts by task-specific and fixed instructions and templates according to the user-requested demonstration lengths. Moreover, we recommend 4 metrics (\textit{true label likelihood}, \textit{Accuracy}, \textit{Macro-F1}, and \textit{ECE}${}_1$~\citep{ECE1}) and encourage users to primarily report results in Macro-F1. By such a methodology, the resulting inconsistency caused by the dataset selection, data sampling, and prompt can be unified. Meanwhile, the descriptive statistics can be simplified by our fixed demonstration sampling, since the experiment repetitions can be reduced.

\subsection{Related Works} 

The models and datasets mentioned in this paper are cited in Appendix~\ref{appendix:model_data_cite}.

\textbf{In-context learning.} Discovered by~\cite{radford2019language}, in-context learning is an emerging few-shot learning paradigm using only feed-forward calculation on LMs. Previous researches are focused on \textbf{improving ICL performance} by calibrating or recalculating the predicted likelihood~\citep{zhao2021calibrate, han2023prototypical, jiang2023generative, fei2023mitigating, zhou2024batch, cho2024token, xu2023knn, abbas2024enhancing, min2022noisy}, aligning LMs towards ICL objective~\citep{wei2023symbol, min2022metaicl, wei2022finetuned, iyer2022opt, bruneticl}, finding better demonstration examples~\citep{liu2022makes, mavromatis2023examples, gonen2023demystifying, qin2023context, van2024context}, and applying better demonstration orders~\citep{lu2022fantastically, liu2024let}. Also, some works try to find the \textbf{principle of ICL} theroically~\citep{olsson2022context, wies2024learnability, huang2023context, collins2024context, jeon2024information} and empirically~\citep{min2022rethinking, yoo2022ground, dai2023can, han2023understanding, wang2023label, kossen2024context, pan2023context, li2024language, cho2024revisiting}. These works typically require benchmarks to validate their algorithms' effectiveness or measure ICL-related data. However, we will show that the selection and usage of these benchmarks are inconsistent, making it difficult to compare and summarize their results across the literature.

\textbf{Currently used ICL benchmarks.} Currently used structured benchmarks for ICL include: BIG-Bench~\citep{srivastava2023beyond} and BIG-Bench Hard~\citep{suzgun2023challenging} for large-scaled and complex tasks, and OPT-IML~\citep{iyer2022opt} for out-of-distribution tasks. However, for general classification tasks, which are quite basic and important for performance validation and principle explanation, no unified benchmarks are found currently, while only piecemeal supervised datasets, such as some subsets of GLUE~\citep{wang2019glue} are used.

\section{Preparation}

\begin{figure}[t]
% \begin{minipage}{0.27\linewidth}
%         \centering
% 		\includegraphics[width=0.99\columnwidth]{Figure/Motivation2SST2.pdf}
% \end{minipage}
% \begin{minipage}{0.27\linewidth}
%         \centering
% 		\includegraphics[width=0.99\columnwidth]{Figure/Motivation2RTE.pdf}
% \end{minipage}\hfill
% \begin{minipage}{0.21\linewidth}
% \centering
%         \scriptsize{\textbf{Acc. of GPT2-XL on SST-2}}\vspace{0.4em}
%         \input{Table/Table_in_Fig1_SST2}
% \end{minipage} 
% \begin{minipage}{0.21\linewidth}
% \centering
%         \scriptsize{\textbf{Acc. of GPT2-XL on RTE}}\vspace{0.4em}
%         \input{Table/Table_in_Fig1_RTE}
% \end{minipage}

% \vspace{0.4em}

\begin{minipage}{0.235\linewidth}
        \centering
		\includegraphics[width=0.99\columnwidth]{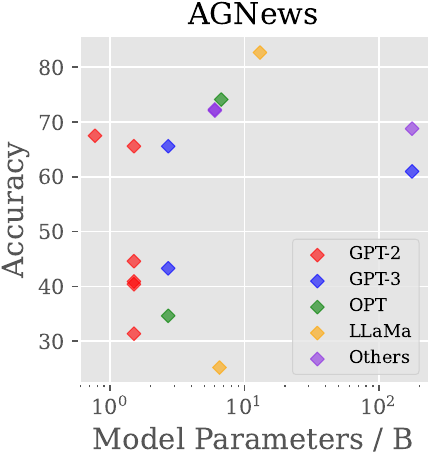}
\end{minipage}
\begin{minipage}{0.235\linewidth}
        \centering
		\includegraphics[width=0.99\columnwidth]{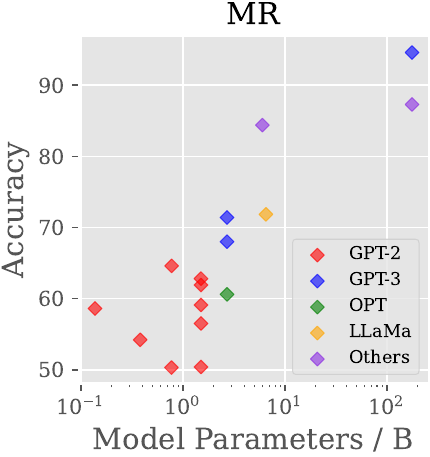}
\end{minipage}\hfill
\begin{minipage}{0.22\linewidth}
\centering
        \scriptsize{\textbf{AGNews}}\vspace{0.4em}
        \resizebox{\linewidth}{!}{
\begin{NiceTabular}{cc}
\toprule \rowcolor{tablebg}
\textbf{Literature} & \textbf{Acc.} \\ \midrule
\citeauthor{gu2023pre} & \textbf{65.6} \\
\citeauthor{han2023prototypical} & 40.9 \\
\citeauthor{zhao2021calibrate} & 44.6 \\
\citeauthor{cho2024token} & \textbf{31.7} \\
\citeauthor{abbas2024enhancing} & 40.4\\ \midrule
\textbf{Range} & \textbf{34.3}\\
\bottomrule
\end{NiceTabular}
}
\end{minipage} 
\begin{minipage}{0.215\linewidth}
\centering
        \scriptsize{\textbf{MR}}\vspace{0.4em}
        \resizebox{\linewidth}{!}{
\begin{NiceTabular}{cc}
\toprule \rowcolor{tablebg}
\textbf{Literature} & \textbf{Acc.} \\ \midrule
\citeauthor{gu2023pre} & 61.9 \\
\citeauthor{han2023prototypical} & 56.5 \\
\citeauthor{cho2024token} & \textbf{50.4} \\
\citeauthor{lu2022fantastically} & 59.1 \\
\citeauthor{xu2023knn} & \textbf{62.8}\\ \midrule
\textbf{Range} & \textbf{12.4}\\
\bottomrule
\end{NiceTabular}
}
\end{minipage}  
\vspace*{-0.6\baselineskip}
\caption{\small Summaries of experimental results in the literature: ICL results are disjoint in the literature. \textbf{Left}: Accuracies of various models with different sizes on specified datasets, where the results do not comply with the scaling laws w.r.t.\ the model scaling. \textbf{Right}: Accuracies of the same model (GPT2-XL) on vanilla ICL inference from different papers, where results have a considerable range even on the same model.}
\label{fig:result_disjoint}
\vspace*{-0.7\baselineskip}
\end{figure}

\subsection{Task Definition: In-context Classification}
\label{sec:task_definintion_ICC}

Typical ICL receives structured input-label pair prompt made from a few-shot \textit{demonstration set} $\mathcal{D}_k = [(x_{1}, y_{1}), (x_{2}, y_{2}), \dots, (x_{k}, y_{k})]$ and a \textit{query} $x_q$ by a \textit{template} $\mathcal{T}$. Fed with the formed input $\mathcal{T}(\mathcal{D}_k, x_q)$, the causal language model $\Theta$ produces a probability distribution of the subsequent token, and in \textbf{I}n-\textbf{c}ontext \textbf{C}lassification (ICC) task with a finite and certain label token set $\mathbb{Y}$ depending on the verbalizer in the prompt template\footnote{Notice that typically each label corresponds to \textbf{one} token.}, we usually select the probabilities of the tokens (of amount $\vert\mathbb{Y}\vert$) presented in the label token set as the output label probabilities. Also, some calibrations $\mathcal{C}$ are used to re-scale the output label probabilities. That is, the output of ICC is:
\begin{equation}
\label{eq:1}
    \mathbf{o}(x_q) = \mathcal{C}\left[\mathrm{TokenSelect}\left(\Theta(\mathcal{T}(\mathcal{D}_k,x_q))\right)\right],
\end{equation}
which is an $\vert\mathbb{Y}\vert$-dimensional probability vector. Obviously, in ICC experiments, we should consider and control all the variables to get comparable results. The $\mathcal{C}$, $\Theta$ are often well-controlled since they are deemed remarkable while the $\mathcal{T}(\cdot, \cdot)$ and $\mathcal{D}_k$ are often overlooked even with an over-expected impact as investigated follows. 

\subsection{Motivation: Investigating the Inconsistency in ICC Experiments}
\label{sec:motivation}

\setlength{\intextsep}{-1pt}
\begin{wrapfigure}[12]{r}{0.35\textwidth}
    \centering
    \includegraphics[width=0.35\textwidth]{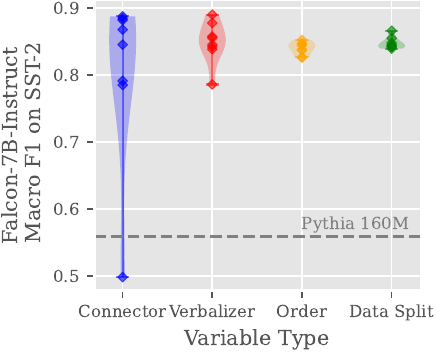}
    \vspace*{-1.6\baselineskip}
    \caption{\small The influence of various prompt variables on ICL. Details are shown in Appendix~\ref{appendix:exp_detail}.}
    \label{fig:variables_analysis}
\end{wrapfigure}

This section claims our proposition: the inconsistency in experimental settings leading to inconsistent results hinders comparison and meta-analysis across the literature. We analyze how the inconsistency happens so that we can propose principles for our benchmark to reduce these inconsistencies.

\textbf{The disjoint in dataset selection.} To find how the dataset usage and experiment implementations are disjoint, we investigated 29 papers that conducted experiments on ICC. We summarize the usage of the most-used 13 datasets and their metrics in Table~\ref{tab:dataset_usage}, where we find that even if some datasets are favored, the datasets used in ICC-related papers are severely inconsistent from an overall perspective. Also, the metrics in these papers are inconsistent.

\textbf{The disjoint in numerical results.} We also investigate the detailed experimental results on the dataset overlaps on these papers in Fig.~\ref{fig:result_disjoint}, where we can find that: \textbf{(1)} These results do not fit the scaling laws~\citep{kaplan2020scaling}, and \textbf{(2)} Results on the same dataset and model are significantly inconsistent. For such a reason, researchers can hardly conduct a robust comparison or meta-analysis between these results reported by different papers.

As an attribution for the numerical disjoint, we find that the consistency in the input formatting is often overlooked in these papers as trivial experimental settings. So, we analyze and measure the impact on various aspects in the input forming in Fig.~\ref{fig:variables_analysis}, where we can see: \textbf{Prompt connectors}, the formative tokens to indicate and connect parts in the prompt (such as the ``Label: '' in Fig.~\ref{fig:processing}), have the most influence on the performance to obscure the difference with a significantly small model, also shown by previous works~\citep{han2023prototypical, voronov2024mind}. Since the prompt templates are quite different across papers, we attribute it to a main reason for inconsistent results in the literature. \textbf{Verbalizers} (label token selection) also strongly influence the results~\citep{min2022rethinking, yoo2022ground} but within our trials, it is not sufficient for an obscure. Random \textbf{Demonstration Order} and \textbf{Data Split} also have a minor influence. It is obvious that restricting these experiment settings can help to obtain stable-over-trial and reproducible results.

\section{\M}

In this section, we propose a \textbf{Sta}ndardized Evaluation Toolkit for \textbf{I}n-\textbf{c}ontext \textbf{C}lassification (\M) to address the disjointing problems mentioned before. Based on the above discussion, we advocate controlling variables in ICL practice, specifically determining a universal testing script with unified datasets and stable-over-trial inputs, to stably test ICL abilities in a consistent condition.

\subsection{Methdology} % 限定什么内容？

\begin{figure}
    \centering
    \includegraphics[width=0.9\linewidth]{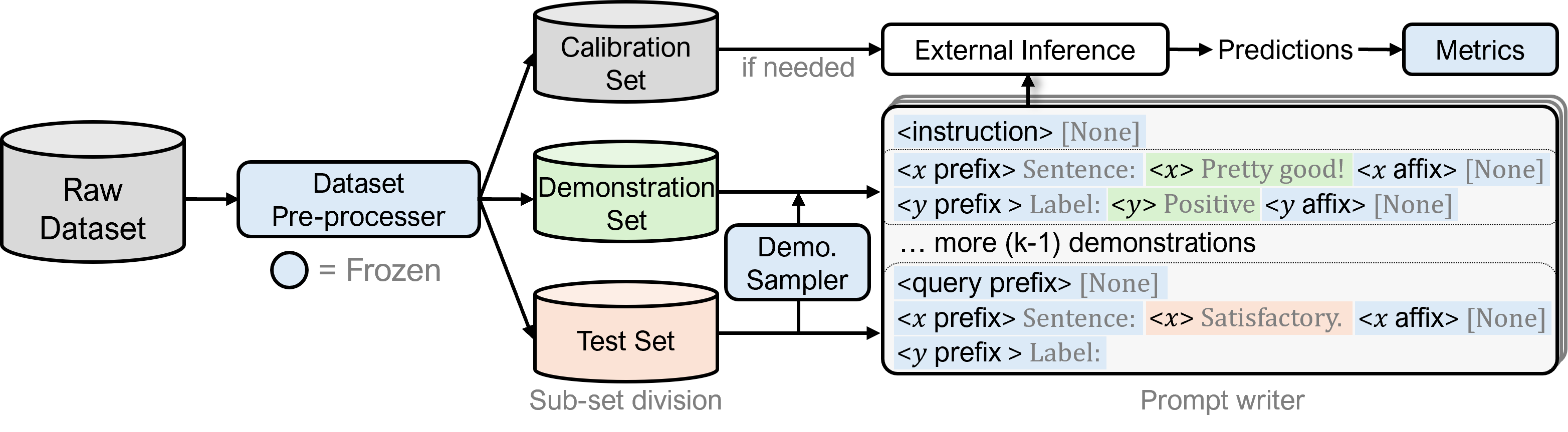}
    \vspace*{-0.4\baselineskip}
    \caption{\small Major schematic diagram of data pre-processing and input forming in~\M. Raw datasets are first divided into 3 sub-sets with a frozen pre-processer for calibration, demonstration, and query. Based on these sub-sets, ICL inputs are built with a frozen demonstration sampler, and a frozen prompt template under a meta-template, shown in gray for an example.}
    \label{fig:processing}
    \vspace*{-0.25\baselineskip}
\end{figure}

\textbf{General principles.} Given the discussion before, we propose an evaluation toolkit~\M~with 10 chosen datasets (listed in Appendix~\ref{appendix:model_data_cite}). For each dataset, we apply a fixed prompt template, data sampling, and demonstration order as the default setting to produce inputs for testing. That is, we ensure that give the $k$ fixed,  the inputs for the test are invariable among different tries. To support the efforts on improving some of the aspects of the ICC process shown in Eq.~\ref{eq:1}, we open all the interfaces to modify all the conditions (including the remarkable ones and the trivial ones, e.g., the prompt template) arbitrarily while the other aspects are recommended to be kept as default. In addition, we try our best to maximize usability and simplicity, so that users only need to call the interface to modify their input forming or reload a forward propagation function (or use our template) to quickly validate their method.

\textbf{Controling variables in assembling ICL inputs.} In detail, as shown in Fig.~\ref{fig:processing}, to build a variable-controlled set of ICL-styled inputs, given each chosen raw supervised classification dataset (Appendix~\ref{appendix:model_data_cite}), we process them as described following: (1) filter some data out from these datasets (typically some over-length and error data), (2) divide the cleaned dataset into 3 splits for tuning additional improvement methods (\textit{calibration set}), demonstration forming (\textit{demonstration set}), and query (\textit{test set}), then, (3) a dataset-specific default input formatting, including a template, a verbalizer, and a fixed demonstration sequence for each query sample in the test set is assigned. As a result, stable-over-trial testing inputs are built based on these default settings. We guarantee the repeatability invariance of this process, so that, dataset selection differences and irrelevant experimental variables can be robustly controlled, increasing the comparability between different implementations, and the variance among experiment repetitions is also reduced to $0$, mitigating the cost of describing statistics and parallel repetition of experiments among different random seeds. 

\textbf{Prompt templates.} As the default setting, we use simple prompt templates as shown as a case in Fig.~\ref{fig:processing} and listed in Appendix~\ref{appendix:prompt_templates}. Also, to enable development on improving the template, we frame these templates into a \textit{meta-template} shown in Fig.~\ref{fig:processing}, where users can freely set the attributes of the meta-template to customize the templates they need. 

\textbf{Is a simple prompt template the optimal choice?} Given that we limit the template to the simplest ones, a doubt is: \textbf{why is such a template representative?} or, \textbf{can an input template faithfully measure the investigated features in a generalizable way?} A previous work~\citep{voronov2024mind} pointed out that some prompt templates can obscure the gap between various models and ICL methods, causing discriminability loss in the testing. Also, the \textit{optimal} or \textit{most faithful} templates are non-transferable between different models or methods, which makes determining \textbf{the most} generalizable template for a benchmark an ill-defined task. Therefore, we follow Occam's Razor, using the simplest but valid template, which also acts as a baseline to support the comparison of research on improving prompt templates. Imageably, such discriminatory suspicion can occur in various aspects of the benchmark designing (such as the sampling, datasets selection, etc., ``why $\dots$ can be used for a faithful test?''), so, to empirically confirm the discriminability of \M, we will calibrate it through scaling laws, that is, if we find larger models perform better on \M, we can preliminarily confirm its discriminability. Also, we ensure the unbiasedness of the dataset sampling (Appendix~\ref{appendix:exp_detail}), and provide sufficient test examples to avoid accidental results. 

% Our answer is: in the scenario of benchmark-making, an experimental setting that reflects \textit{all} ``essential elements'' consistently and faithfully is ill-defined. Specifically, if a method is asserted effective, it should be effective in any valid experimental setting. In such a scenario, detailed experimental setups (such as the prompt template) are trivial to asserting the effectiveness of methods or conclusions to be tested, even if they numerically influence the experimental results. That's also why most of the results in the literature are relatively reliable, even if their settings and results cannot be compared crosswise. Therefore, based on Occam's Razor principle, we use the simplest template, which also acts as the baseline to support research on improving prompt templates [cite]. While, we ensure the unbiasedness of the dataset slicing [Appendix], and provide sufficient test cases to avoid accidental results.

Specifically, we propose several sub-benchmarks for different purposes based on the above principle, which is introduced below.

\subsection{\MN: Basic Standardized Benchmark for ICC}

\setlength{\intextsep}{-1pt}
\begin{wrapfigure}[12]{r}{0.30\textwidth}
    \centering
    \includegraphics[width=0.30\textwidth]{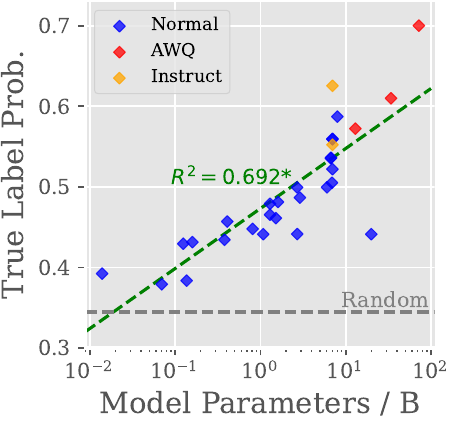}
    \vspace*{-1.5\baselineskip}
    \caption{\small TLP results against model parameter numbers.}
    \label{fig:basic_res}
\end{wrapfigure}

As the main objective of this paper, to test the basic prediction capacity of an ICC inference $\mathbf{o}(x_q)$, we build~\MN~with standard input form and typical metrics for classification tasks.

\textbf{Metric.} We apply 4 metrics for this sub-benchmark: (1) Accuracy, (2) \textbf{T}rue \textbf{l}abel \textbf{p}robability (TLP), (3) Macro F1, (4) \textbf{E}xpected \textbf{c}alibration \textbf{e}rror\textbf{-1} (ECE-1)~\citep{guo2017calibration, naeini2015obtaining}, with various linearity~\citep{schaeffer2024emergent} and application focus. For each test sample $x_i$ in the test set $\{x_i\}_{i=1}^N$, given the ground-truth label $y_i$ and the prediction output $\mathbf{o}\left(x_i\right)$, the metrics can be formulated as below: 
\begin{equation*}
    \text{Acc} = \frac{1}{N} \sum_{i=1}^N \mathbbm{1}\left[\mathop{\mathrm{argmax}}\limits_{\hat{y}\in \mathbb{Y}}\mathbf{o}\left(x_q\right) = y_i\right],\quad\text{TLP} = \frac{1}{N} \sum_{i=1}^N \mathbf{o}(x_i)_{y_i}, \quad \quad
\end{equation*}
\begin{equation*}
    \text{Macro F1} = \frac{1}{\vert\mathbb{Y}\vert} \sum_{j=1}^{\vert\mathbb{Y}\vert} \frac{2 \cdot \text{Precision}_j \cdot \text{Recall}_j}{\text{Precision}_j + \text{Recall}_j}, \quad \text{Precision}_j = \frac{\text{TP}_j}{\text{TP}_j + \text{FP}_j}, \quad \text{Recall}_j = \frac{\text{TP}_j}{\text{TP}_j + \text{FN}_j},\quad\quad
\end{equation*}
\begin{equation*}
    \text{ECE}_1 = \sum_{b=1}^B \frac{|B_b|}{N} \left| \text{acc}(B_b) - \frac{1}{|B_b|} \sum_{i \in B_b} \max_j \mathbf{o}(x_i)_j \right|,\quad \text{acc}(B_b) = \frac{1}{|B_b|} \sum_{i \in B_b} \mathbbm{1}\left[\mathop{\mathrm{argmax}}\limits_{\hat{y}\in \mathbb{Y}}\mathbf{o}\left(x_q\right) = y_i\right], \quad\quad
\end{equation*}

where $\text{TP}_j$, $\text{FP}_j$, and $\text{FN}_j$ are the true positive, false positive, and false negative counts for class $j$, respectively, $B_b$ is the set of prediction samples in the $b$-th confidence bin within a total $B$ bins (set to $10$ in default), that is, the prediction with max value in $\left[\frac{b-1}{B}, \frac{b}{B}\right)$, $|B_b|$ is the number of samples in $B_b$. For each dataset, we test the ICC inference $\mathbf{o}(x_q)$ on these metrics, and take the averaged result on the 10 datasets.

\textbf{Benchmark results: ICC baselines and scaling laws.} We test 29 modern LMs on \MN, with TLP results shown in Fig.~\ref{fig:basic_res}, where a clear log-linear scaling law can be observed. This outcome remains valid even amid concerns about the discriminability of a simple prompt template, as the observed scaling law is widely recognized~\citep{kaplan2020scaling, bahri2024explaining} as a first principle in neural language models, and suggests that our approach remains robust.

\begin{figure}
    \centering
    \includegraphics[width=1\linewidth]{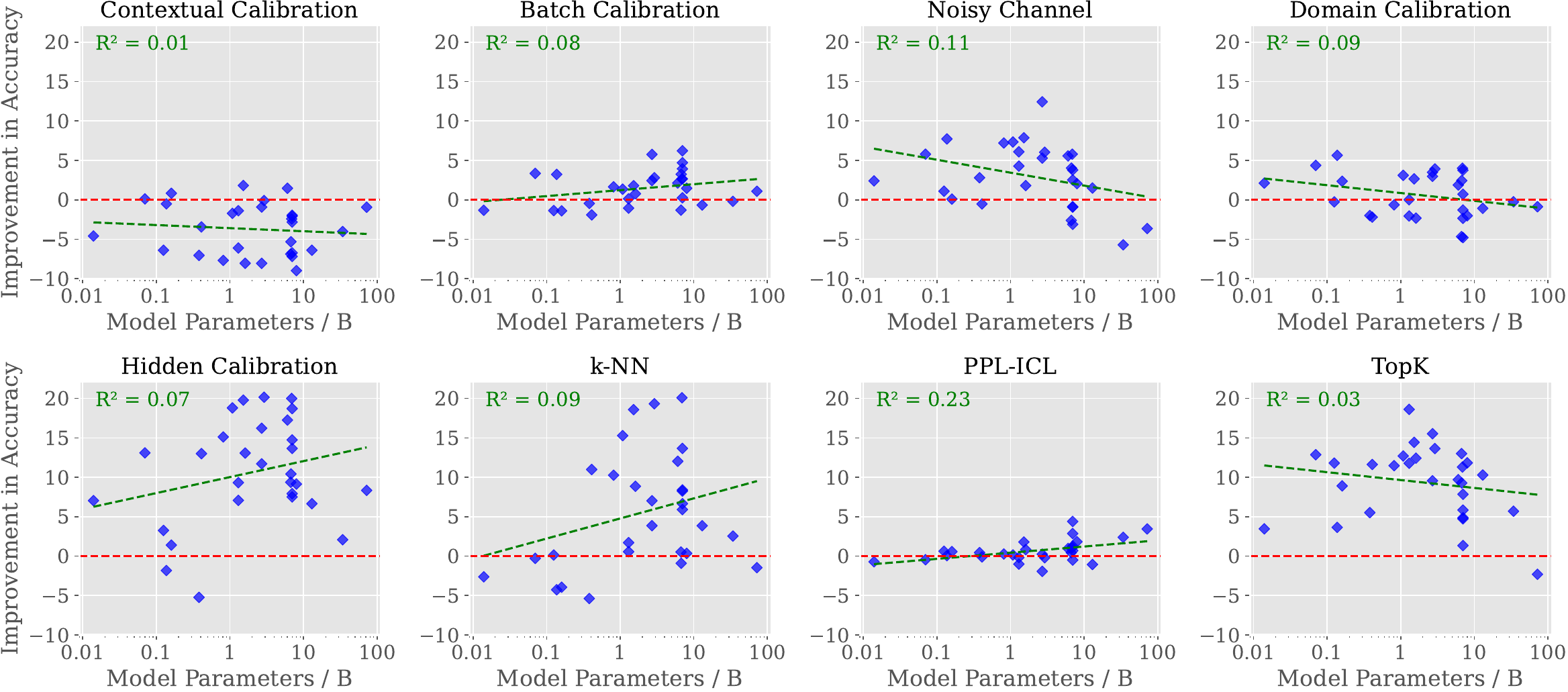}
    \vspace*{-1.5\baselineskip}
    \caption{\small Absolute accuracy (\%) improvement of various ICL-improving methods against model parameter numbers. }
    \label{fig:improve_res}
    \vspace*{-0\baselineskip}
\end{figure}

% Please add the following required packages to your document preamble:
% \usepackage{booktabs}
\begin{table}[]
\setlength{\tabcolsep}{5pt}
\renewcommand{\arraystretch}{1.0}
\setlength{\abovecaptionskip}{2pt}
\centering
\caption{\small Accuracies (\%) of 10 representative models and 10 inference methods on \MN. The best inference method on each model is shown in \textbf{bold}, *: Instruction-tuned models. More detailed results are in Appendix~\ref{appendix:more_res}.}
\label{table:acc_improving}
\resizebox{\linewidth}{!}{
\begin{NiceTabular}{ccccccccccc}
\toprule
\rowcolor{tablebg}
\textbf{Models} &
  \textbf{GPT-2} &
  \textbf{GPT-2} &
  \textbf{OPT} &
  \textbf{Llama2} &
  \textbf{Qwen2*} &
  \textbf{Qwen2} &
  \textbf{Llama3} &
  \textbf{Llama2} &
  \textbf{Llama2} &
  \textbf{Qwen2*} \\
  \rowcolor{tablebg}
Para.\ \# / B              & 0.14 & 1.6  & 2.7   & 6.7  & 7.6  & 7.6  & 8.0  & 13$_{\mathrm{AWQ}}$ & 34$_{\mathrm{AWQ}}$ & 72$_{\mathrm{AWQ}}$  \\ \midrule
\multicolumn{11}{c}{\textit{-- Basic Inference --}}  \\ 
\textbf{Vanilla}               & 40.24 & 50.31 & 51.48 & 57.32 & 66.87 & 61.19 & 64.42 & 63.47 & 66.83 & 72.36 \\
\textbf{Noisy Ch.}              & \textbf{47.97} & 52.12 & 56.78 & 61.35 & 63.76 & 63.71 & 66.47 & 64.96 & 61.11 & 68.72 \\ \midrule
\multicolumn{11}{c}{\textit{-- Output Probability Calibration --}}   \\ 
\textbf{Context.\ Ca.}            & 39.73 & 42.27 & 43.43 & 52.00 & 64.04 & 59.23 & 55.44 & 57.07 & 62.80 & 71.43 \\
\textbf{Domain Ca.}              & 45.89 & 47.99 & 54.46 & 59.77 & 65.58 & 61.90 & 62.40 & 62.38 & 66.57 & 71.48 \\
\textbf{Batch Ca.}               & 43.47 & 51.08 & 53.93 & 60.57 & 69.52 & 65.90 & 65.89 & 62.80 & 66.63 & 73.45 \\ \midrule
\multicolumn{11}{c}{\textit{-- Input Reforming --}}   \\ 
\textbf{PPL-ICL}                & 40.30 & 51.13 & 51.73 & 58.34 & 69.73 & 65.59 & 66.25 & 62.40 & 69.21 & 75.81 \\
\textbf{TopK}                   & 43.89 & 62.73 & 67.01 & 66.60 & 68.20 & 67.03 & 76.25 & 73.75 & 72.52 & 70.06 \\
\textbf{SA-ICL}                 & 43.85 & 63.28 & \textbf{69.24} & \textbf{68.18} & 69.34 & 67.54 & \textbf{77.13} & \textbf{75.37} & \textbf{74.34} & 70.78 \\ \midrule
\multicolumn{11}{c}{\textit{-- Downstream Classifier --}}   \\ 
\textbf{$k$-NN}                 & 35.98 & 56.42 & 59.17 & 55.34 & 73.52 & 67.10 & 67.32 & 69.26 & 70.90 & 64.76  \\
\textbf{Hidden Ca.}              & 38.42 & \textbf{63.39} & 63.19 & 67.74 & \textbf{80.53} & \textbf{75.94} & 73.56 & 70.12 & 68.91 & \textbf{80.70} \\
\bottomrule
\end{NiceTabular}}
\vspace*{-1\baselineskip}
\end{table}

\textbf{Benchmark results: on the ICL-improving methods.} Moreover, we evaluate 9 ICL-improving methods (referenced in Appendix~\ref{appendix:model_data_cite}) on \MN, with ensuring to the best of our ability that the amount of additional data (if used) remains consistent across methods for a fair comparison. Table~\ref{table:acc_improving} presents the results for selected representative models, where Hidden Calibration~\citep{cho2024token}, and SA-ICL~\citep{wu2023self} outperforms frequently among these models. Additionally, scaling laws are not observed on the \textit{improvement} of accuracies by these ICL-improving methods, as shown in Fig.~\ref{fig:improve_res}.

\subsection{\MD: Diagnostic Evaluation for ICC}

To enrich the usage of \M, we propose \MD~for a set of fine-grained diagnostic analyses, focusing on two key aspects: (1) \textbf{prediction bias}, which identifies tendencies of LMs toward specific label tokens, and (2) \textbf{prediction robustness}, which evaluates the robustness of LMs against subtle variations in ICL inputs. \MD~is composed of several sub-tasks, each for an aspect of the diagnostic, and utilizes various but stable-over-trial input forms different from the \M, to be introduced in detail as follows.

\begin{figure}
    \centering
    \includegraphics[width=0.95\linewidth]{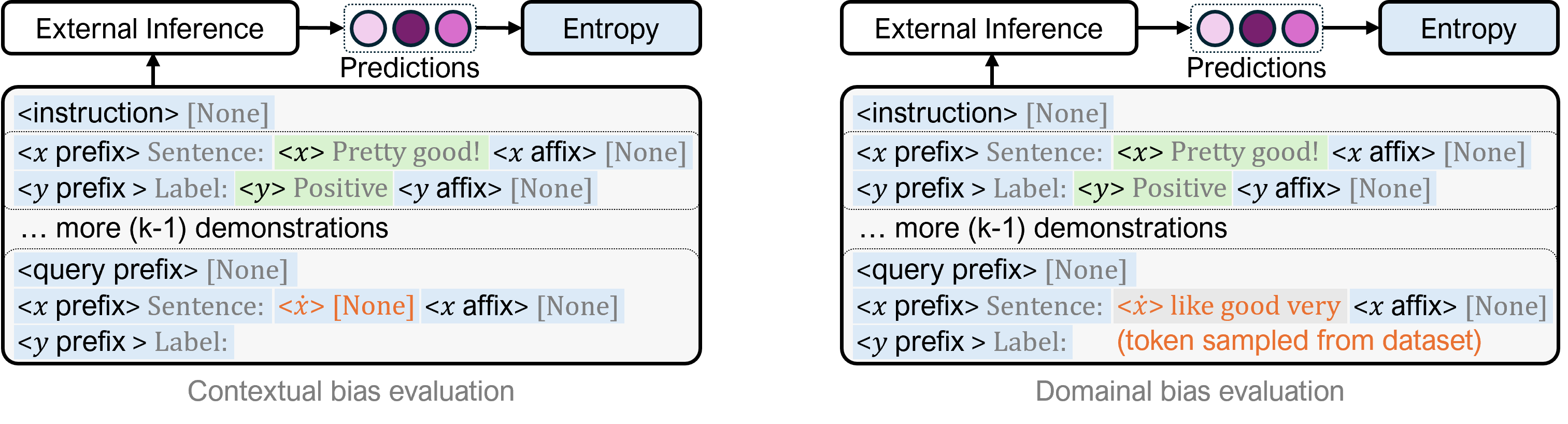}
    \vspace*{-0.8\baselineskip}
    \caption{\small Evaluation process for (\textbf{left}) contextual bias, where \textcolor[HTML]{e97132}{empty pseudo queries ($\dot{x}$)} are included in the input; and (\textbf{right}) domainal bias, where \textcolor[HTML]{e97132}{pseudo queries ($\dot{x}$)} consist of token sequences i.i.d.\ sampled from the token distribution of all real instances $x$. Entropy is calculated on the output distribution to measure how biased the distribution is.}
    \label{fig:bias_processing}
    \vspace*{-0\baselineskip}
\end{figure}

\textbf{Prediction bias evaluation.} We propose evaluations for 3 types of prediction bias: 

(1) \textbf{Contextual bias}~\citep{zhao2021calibrate}\textbf{.} Given input with a meanless pseudo query $\dot{x}$ (in the calculation of contextual bias, an empty query is used as this $\dot{x}$, as shown in Fig.~\ref{fig:bias_processing}), an ideal non-bias inference is expected to produce neutral prediction with averaged probabilities\footnote{e.g., $\left[\frac{1}{3}, \frac{1}{3}, \frac{1}{3}\right]$ for a 3-way scenario.}, producing the maximum entropy on the prediction distribution. Therefore, we use an entropy-related metric to evaluate the bias magnitude:
\begin{equation}
    \text{Contextual Bias} = -\frac{1}{N}\sum_{i=1}^N \mathrm{H}\left[\mathbf{o}\left(\dot{x}\right)\right],
\end{equation}
where the $N$ is the input instance number same as the \MN\footnote{In the implementation, we replaced the test set with pseudo samples, keeping the input forming processing as before.}, and $\mathrm{H}[\cdot]$ is the entropy calculation. Notice that we use a minus sign to ensure that a larger value corresponds to a stronger bias. 

(2) \textbf{Domainal bias}~\citep{fei2023mitigating}\textbf{.} Following the same principle as the contextual bias, the sampling method for $\dot{x}$ is changed to randomly sample a sequence of tokens under the frequency distribution of the $x$s from the calibration set to calculate the domainal bias. 

(3) \textbf{Empirical bias.} Contextual bias and domainal bias offer observations on the prior or ``nature'' output tendency without the influence of query. However, pseudo queries in the inputs of the aforementioned 2 bias calculations can cause a distribution gap between the bias evaluation and practical scenario, potentially harming their applicability. So, different from the above 2 bias evaluations, empirical bias utilizes real test inputs same with \MN, and collects the averaged output probability distributions to calculate KL divergence with the ground-truth label frequency counted from the test set:
\begin{equation}
    \text{Empirical Bias} = \mathrm{D}_{\text{KL}} \left( \frac{1}{N} \sum_{i=1}^N \mathbf{o}(x_i) \, \Bigg\Vert \, \mathbf{p}_{\text{true}} \right),\quad \mathbf{p}_{\text{true}, j} = \frac{\mathop{\mathrm{count}}(y_i=j)}{N}.
\end{equation}

\vspace{0.3\baselineskip}
\textbf{Prediction robustness evaluation.} We propose evaluations for 3 types of prediction robustness: 

(1) \textbf{Prompt template robustness.} We construct 9 prompt templates under the meta-template shown in Fig.~\ref{fig:processing} (detailed in Appendix~\ref{appendix:more_prompt_temp}) and repeat the standard inference process as \MN~once for each template, totaling 9 repeats. Then, for each demonstrations-query sample, we evaluate the prediction consistency among the 9 parallel repeats, by calculating the majority rate of predicted label tokens\footnote{That is, the prediction after the $\mathrm{argmax}$ operation on the predicted probability distribution.}, regardless the correctness. For example, if 6 ``negative'' prediction is given to one query over 9 templates, then the consistency is calculated as $\frac{6}{9}$. We take the average of consistency on all the demonstrations-query instances as the prompt template robustness evaluation.

\begin{figure}
    \centering
    \includegraphics[width=\linewidth]{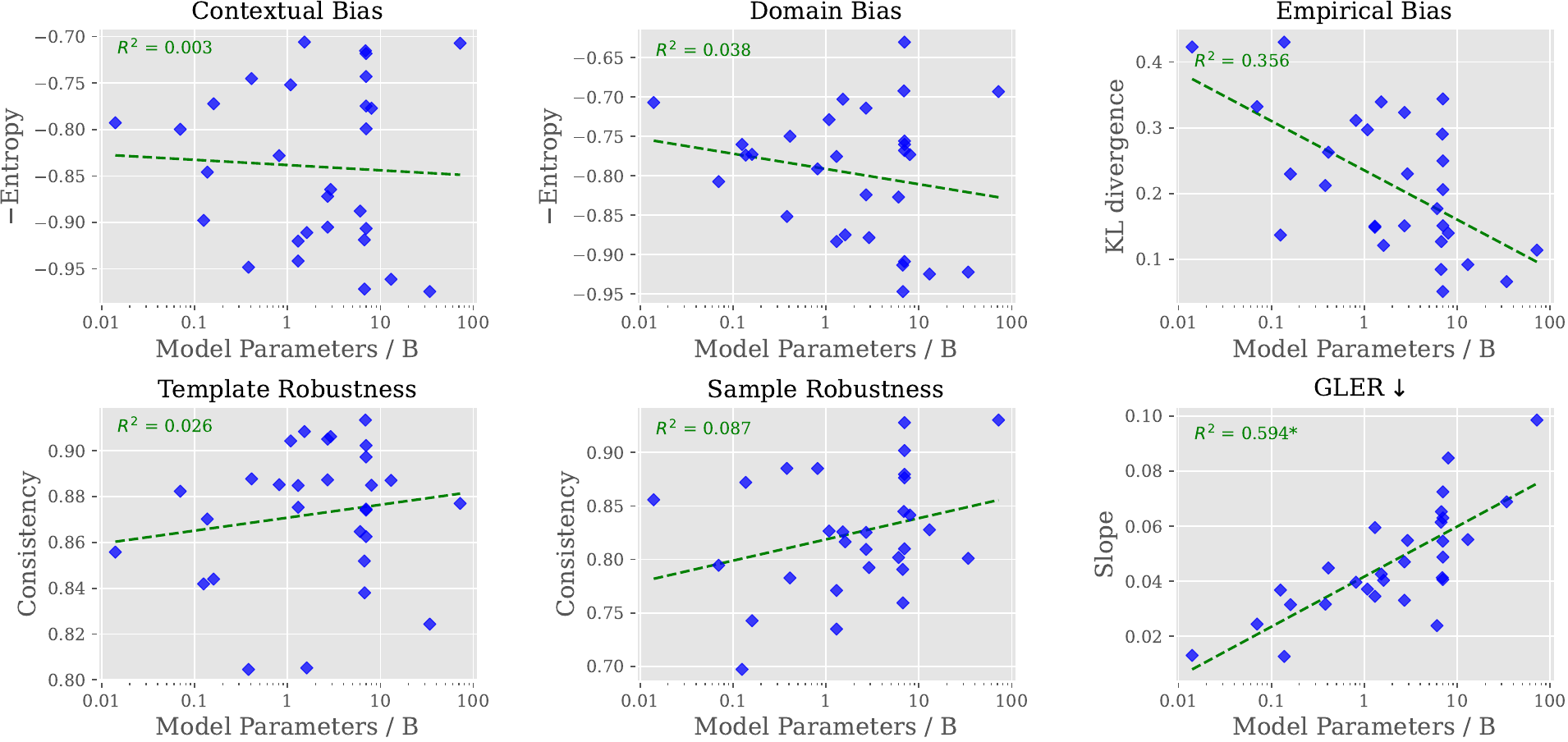}
    \vspace*{-1.5\baselineskip}
    \caption{\small Diagnostic results against model parameter numbers.}
    \label{fig:diag_res}
    \vspace*{-0.8\baselineskip}
\end{figure}

% Please add the following required packages to your document preamble:
% \usepackage{booktabs}
\begin{table}[]
\setlength{\tabcolsep}{5pt}
\renewcommand{\arraystretch}{1.0}
\setlength{\abovecaptionskip}{2pt}
\centering
\caption{\small Diagnostic results of 10 representative models on \MD. *: Instruction-tuned models. More detailed results are in Appendix~\ref{appendix:more_res}.}
\label{table:diag}
\resizebox{\linewidth}{!}{
\begin{NiceTabular}{cccccccccccc}
\toprule
\rowcolor{tablebg}
\textbf{Models}  &  & \textbf{GPT-2} & \textbf{GPT-2} & \textbf{OPT} & \textbf{Llama2} & \textbf{Qwen2*} & \textbf{Qwen2} & \textbf{Llama3} & \textbf{Llama2} & \textbf{Llama2} & \textbf{Qwen2*} \\ \rowcolor{tablebg}
Para. \# / B  & \multirow{-2}{*}{\textbf{Metric}} & 0.14 & 1.6  & 2.7& 6.7& 7.6& 7.6  & 8.0 & 13$_\mathrm{AWQ}$& 34$_\mathrm{AWQ}$& 72$_\mathrm{AWQ}$\\ \midrule
\textbf{Context.\ Bias}  & $-$Entropy & -0.85 & -0.91 & -0.90  & -0.97  & -0.74  & -0.77 & -0.78  & -0.96  & -0.97 & -0.71 \\
\textbf{Domain Bias}& $-$Entropy  & -0.77 & -0.87 & -0.82  & -0.95  & -0.76  & -0.76 & -0.77  & -0.92  & -0.92 & -0.69 \\
\textbf{Empirical Bias}& KL div.  & 0.43 & 0.12 & 0.15  & 0.13  & 0.15  & 0.25 & 0.14  & 0.09  & 0.07 & 0.11 \\ \midrule
\textbf{Temp.\ Robust.} & Cons.(\%)&  87.0 &  80.5  &  88.7  &  83.8  &  87.4  &  89.7  &  88.5  &  88.7  &  82.4 & 87.7\\
\textbf{Sample Robust.}& Cons.(\%)&  87.2 &  81.6 &  80.9  &  75.9  &  87.6  &  92.8 &  84.2  &  82.8  &  80.1 & 93.0 \\
\textbf{GLER}$(\downarrow$)& Slope$/0.1$& 0.13& 0.32& 0.47 & 0.65 & 0.73 & 0.63& 0.85 & 0.55 & 0.69 & 0.98\\ \bottomrule
\end{NiceTabular}}
\vspace*{-1\baselineskip}
\end{table}

(2) \textbf{Demonstration sampling robustness.} Similarly, we repeat the inference among 8 demonstration sequences sampled randomly and calculate the demonstration sampling robustness by consistency.

(3) \textbf{Label noise robustness: GLER}~\citep{yoo2022ground}\textbf{.} We repeat the inference the same as the \MN~on various label noise rates $p$, which controls the percentage of falsified labels in the demonstrations (for example, given $p=0.5$ and $k=4$, then 2 labels in the demonstrations are turned into wrong labels in the input-forming processing). For 5 uniformly sampled $p$ from $0$ to $1$, we can get 5 accuracies $\mathrm{Acc}_p$, then we run a linear regression on the $\mathrm{Acc}_p$ against $p$, and get a slope, which is used as the metric for the label noise \textit{sensitivity} (also named GLER by~\cite{yoo2022ground}), which is a negative metric for the label noise robustness.

\textbf{Benchmark results.} Same as \MN, we test 29 modern LMs on \MD, with results shown in Fig.~\ref{fig:diag_res} and Table~\ref{table:diag}, where the bias is globally descending, and the robustness is ascending against the model scaling, while with a statistically insignificant correlation strength. However, the GLER has a significant ascending correlation against the model scale (that is, descending robustness), suggesting that a larger LM can be affected by the label noise more severely, aligning with the previous findings~\citep{wei2023larger, shi2024larger, cho2024revisiting}.

\section{Discussion}

\textbf{Summary.} Motivated by the disjoint in the literature on the evaluation of in-context learning, this paper proposes a standardized evaluation toolkit \M~for classification task under the in-context learning paradigm and conducts extensive measurement on various LMs.

\setlength{\intextsep}{-1pt}
\begin{wrapfigure}[22]{r}{0.5\textwidth}
    \centering
    \includegraphics[width=\linewidth]{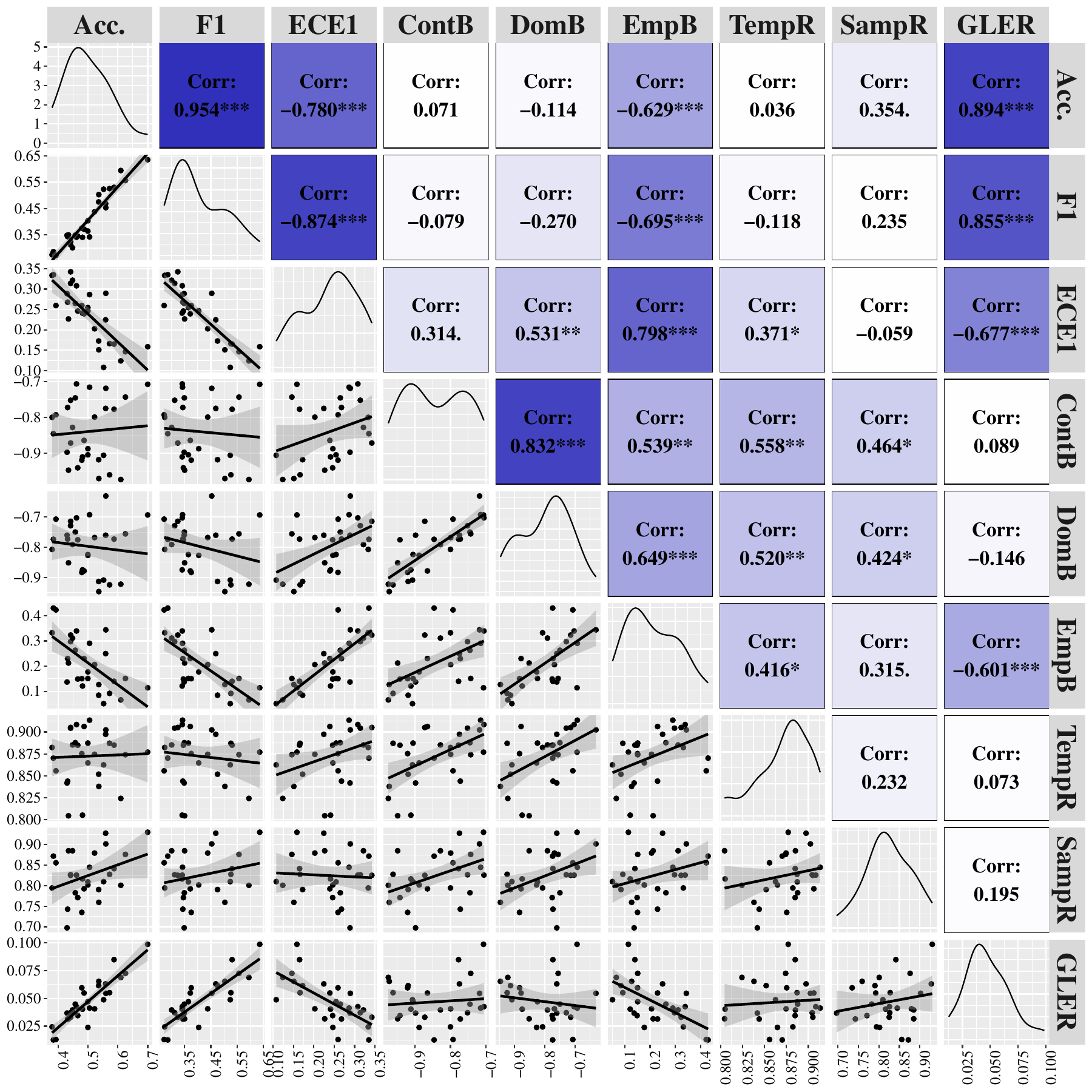}
    \vspace*{-1.9\baselineskip}
    \caption{\small Multivariate covariate analysis on the metrics in this paper. Spearman correlation is calculated.}
    \label{fig:covar}
    \vspace*{-0.8\baselineskip}
\end{wrapfigure}

\textbf{Rethinking the influence of bias and robustness of ICL.} We visualize the statistical covariate among the metrics evaluated in this paper, shown in Fig.~\ref{fig:covar}. We highlight some non-trivial covariation or dis-covariation, as: (1) Weak correlation between accuracy/F1 and contextual/domainal bias suggests a limited influence of prediction bias to ICL inference, aligning with the results in Fig.~\ref{fig:improve_res}, where calibration methods are with relatively weaker improvement on the accuracy, while the correlations from ECE-1 are relatively stronger, suggesting that the benefit of mitigating such bias is towards a more faithful prediction, rather a more accurate one. (2) Template robustness tends to be positively correlated with ECE-1, and sampling robustness tends to be positively correlated with accuracy, indicating the benefits of improving the prediction robustness, (3) A larger GLER (lower robustness) seems to improve accuracy, but given the confounding variable of model size, such a positive covariation can not be regarded as a reliable causal relationship, where further causal inference can be beneficial for a deeper insight.

\textbf{Position of this paper.} The problem addressed in this paper is significant and holds substantial value for future works in the ICL area, while the authors believe the novelty of the works in this paper will not impress readers a lot, with just ``dirty works'' towards an easy-to-use toolkit. Given these considerations, we have chosen to release this work as a scientific report rather than submit it to a peer-reviewed conference. 

\textbf{Limitation.} (1) This paper and \M~mainly take classification tasks in the scope, while other tasks without a fixed and limited output space can still benefit from ICL, raising a need for evaluation of ICL models and methods on these tasks. (2) Also, single-sentence classification tasks are also limited, as tasks with more complex input forms, such as what is defined in RTE~\citep{RTE1, RTE2, RTE3, RTE5} can be non-trivial extensions of the current tasks, and may be considered in the future construction of the benchmark.

\subsubsection*{Content Reuse Authorization}

The author authorizes anyone citing this paper to directly reuse any tables, figures, and text passages from it to assist in explaining their experimental setups to enhance the reproducibility of their work.

\subsubsection*{Acknowledgments}

Mariko Kato and Yoshihiro Sakai participated in our discussion but didn't meet the author criteria.

\bibliography{iclr2025_conference}
\bibliographystyle{tmlr}

\clearpage

\appendix

\section{Models, Datasets, and Inference Methods}
\label{appendix:model_data_cite}

\paragraph{Models} GPT2 (137M, M: 380M, L: 812M, XL: 1.61B)~\citep{radford2019language}, Pythia~\citep{biderman2023pythia} (14M$\sim$6.9B), OPT~\citep{zhang2022opt} (1.3B, 2.7B, 6.7B), GPT-J~\citep{wang2021gpt} (6B), GPT-NEO (125M, 1.3B, 2.7B)~\citep{gpt-neo}, FALCON (7B, 7B instruction-tuned)~\citep{falcon40b}, Qwen (7B)~\citep{qwen}, QWen2 (7B, 7B instruction-tuned, 72B instruction-tuned)~\citep{qwen2}, Llama2 (7B, 13B, 34B)~\citep{touvron2023llama}, and Llama3~\citep{grattafiori2024llama}. In this paper, models larger than 10B are quantified by \textbf{A}ctivation-aware \textbf{W}eight \textbf{Q}uantization (AWQ)~\citep{lin2024awq}.

\paragraph{Datasets} In~\M, we utilize the following datasets: SST-2~\citep{sst2}, MR~\citep{MR}, Financial Phrasebank~\citep{FP}, SST-5~\citep{sst2}, TREC~\citep{trec1, trec2}, AGNews~\citep{Zhang2015CharacterlevelCN}, Subjective~\citep{subj}, Tweet Eval Emotion~\citep{mohammad-etal-2018-semeval}, Tweet Eval Hate~\citep{basile-etal-2019-semeval}, Hate Speech 18~\citep{gibert2018hate}. Also, we refer to additional datasets in this paper, mainly in Table~\ref{tab:dataset_usage}, they are: RTE~\citep{RTE1, RTE2, RTE3, RTE5}, DBPedia~\citep{dbpedia},  MRPC~\citep{mrpc}, CommitmentBank~\citep{commitmentbank}, and WNLI~\citep{levesque2012winograd}. 

\paragraph{Inference methods and their settings.} In this paper, especially Table~\ref{table:acc_improving}, we investigate these inference methods (i.e., ICL-improving methods): Noisy channel~\citep{min2022noisy}, Contextual calibration~\citep{zhao2021calibrate}, Domain calibration~\citep{fei2023mitigating}, Batch calibration~\citep{zhou2024batch}, PPL-ICL~\citep{gonen2023demystifying}, TopK~\citep{liu2022makes}, SA-ICL~\citep{wu2023self}, $k$-NN~\citep{xu2023knn}, and Hidden Calibration~\citep{cho2024token}.

\section{Implementation Details}
\label{appendix:detail}

\subsection{Default \M~Settings}
\label{appendix:prompt_templates}
\label{appendix:more_prompt_temp}

\textbf{Prompt templates.} We substitute the default attributes shown in Table~\ref{tab:template} into the meta-template (Fig.~\ref{fig:processing}) for the default prompt templates.

\begin{table}[t]
\setlength{\tabcolsep}{5pt}
\renewcommand{\arraystretch}{1.0}
\setlength{\abovecaptionskip}{2pt}
\centering
\caption{\small Default prompt template for each dataset under the meta-template.}
\label{tab:template}
\resizebox{1\columnwidth}{!}{
\begin{NiceTabular}{@{}ccc|ccc@{}}
\toprule \rowcolor{tablebg}
\textbf{Dataset}           & \textbf{Attribute}    & \textbf{Value} & \textbf{Dataset} & \textbf{Attribute}    & \textbf{Value} \\ \midrule
\multirow{7}{*}{SST-2} & Instruction  & -           & \multirow{7}{*}{MR}    & Instruction  &  -    \\
                  & $x$ prefix   & ``sentence: ''    &                           & $x$ prefix   & ``revies: '' \\
                  & $x$ affix    &  `` ''           &                           & $x$ affix    &  `` ''  \\
                  & $y$ prefix   &  ``sentiment: ''   &                           & $y$ prefix   & ``sentiment: '' \\
                  & $y$ affix    & ``$\backslash$n''  &                         & $y$ affix    & ``$\backslash$n'' \\
                  & Query prefix &  -               &                           & Query prefix &  -    \\
                  & Label space  & [``positive'', ``negative'']      &          & Label space  & [``positive'', ``negative''] \\ \midrule
\multirow{7}{*}{\thead{Financial\\ Phrasebank}} & Instruction  & -           & \multirow{7}{*}{SST-5}    & Instruction  &  -    \\
                  & $x$ prefix   & ``sentence: ''    &                           & $x$ prefix   & ``sentence: '' \\
                  & $x$ affix    &  `` ''           &                           & $x$ affix    &  `` ''  \\
                  & $y$ prefix   &  ``sentiment: ''   &                           & $y$ prefix   & ``sentiment: '' \\
                  & $y$ affix    & ``$\backslash$n''  &                         & $y$ affix    & ``$\backslash$n'' \\
                  & Query prefix &  -               &                           & Query prefix &  -    \\
                  & Label space  & [``positive'', ``neutral'', ``negative'']      &          & Label space  & [``poor'', ``bad'', ``neutral'', ``good'', ``great''] \\ \midrule
\multirow{7}{*}{\thead{TREC}} & Instruction  & -           & \multirow{7}{*}{AGNews}    & Instruction  &  -    \\
                  & $x$ prefix   & ``question: ''    &                           & $x$ prefix   & ``news: '' \\
                  & $x$ affix    &  `` ''           &                           & $x$ affix    &  `` ''  \\
                  & $y$ prefix   &  ``target: ''   &                           & $y$ prefix   & ``topic: '' \\
                  & $y$ affix    & ``$\backslash$n''  &                         & $y$ affix    & ``$\backslash$n'' \\
                  & Query prefix &  -               &                           & Query prefix &  -    \\
                  & Label space  & \thead{[``short'', ``entity'', ``description'', \\``person'', ``location'', ``number'']}      &          & Label space  & [``world'', ``sports'', ``business'', ``science''] \\ \midrule   
\multirow{7}{*}{\thead{Subjective}} & Instruction  & -    & \multirow{7}{*}{\thead{Tweet\\Eval\\Emotion}} & Instruction  &  -    \\
                  & $x$ prefix   & ``review: ''    &                           & $x$ prefix   & ``tweet: '' \\
                  & $x$ affix    &  `` ''           &                           & $x$ affix    &  `` ''  \\
                  & $y$ prefix   &  ``subjectiveness: ''   &                           & $y$ prefix   & ``emotion: '' \\
                  & $y$ affix    & ``$\backslash$n''  &                         & $y$ affix    & ``$\backslash$n'' \\
                  & Query prefix &  -               &                           & Query prefix &  -    \\
                  & Label space  & [``objective'', ``subjective'']      &          & Label space  & [``anger'', ``joy'', ``positive'', ``sad''] \\ \midrule  
\multirow{7}{*}{\thead{Tweet\\Eval\\Hate}} & Instruction  & -    & \multirow{7}{*}{\thead{Hate\\Speech\\18}} & Instruction  &  -    \\
                  & $x$ prefix   & ``tweet: ''    &                           & $x$ prefix   & ``tweet: '' \\
                  & $x$ affix    &  `` ''           &                           & $x$ affix    &  `` ''  \\
                  & $y$ prefix   &  ``hate speech: ''   &                           & $y$ prefix   & ``hate speech: '' \\
                  & $y$ affix    & ``$\backslash$n''  &                         & $y$ affix    & ``$\backslash$n'' \\
                  & Query prefix &  -               &                           & Query prefix &  -    \\
                  & Label space  & [``normal'', ``hate'']      &          & Label space  & [``normal'', ``hate'', ``skip'', ``relation''] \\
                  \bottomrule
\end{NiceTabular}
}
\vspace*{\baselineskip}
\end{table}

\textbf{Raw dataset divisions.} We divide the raw dataset into trisection with specifications shown in Table~\ref{table:dataset_div}. 

\textbf{Candidate template for the template sensitivity evaluation.} For the attribute ``Instruction'', ``$x$ prefix'', ``$y$ prefix'', and ``$y$ affix'', for each dataset, we provide 2 more alternative options, as shown in Table~\ref{tab:alter}. Then, we apply a $\mathrm{L}9(3^4)$ Taguchi's orthogonal array~\citep{taguchi1987taguchi} to sample a combination of attributes to substitute into the meta-template as the candidate templates.

\begin{table}[]
\setlength{\tabcolsep}{5pt}
\renewcommand{\arraystretch}{1.0}
\setlength{\abovecaptionskip}{2pt}
\centering
\caption{\small Raw dataset division size for each sub-dataset.}
\label{table:dataset_div}
\resizebox{0.7\linewidth}{!}{
\begin{NiceTabular}{ccccccccccc}
\toprule
\rowcolor{tablebg}
& \textbf{SST$_{2}$} & \textbf{MR} & \textbf{FP} & \textbf{SST$_{5}$} & \textbf{TREC} & \textbf{AGN} & \textbf{Subj.} & \textbf{TEE} & \textbf{TEH} & \textbf{HS18} \\ \midrule
\textbf{Calibration Set} & 1024 & 1024 & 1024 & 1024 & 1024 & 1024 & 1024 & 1024 & 1024 & 1024 \\
\textbf{Demonstration Set} & 4096 & 4096 & 512 & 4096 & 4096 & 4096 & 4096 & 4096 & 3192 & 4096 \\
\textbf{Test Set} & 512 & 512 & 512 & 512 & 512 & 512 & 512 & 512 & 512 & 512 \\ \bottomrule
\end{NiceTabular}}
\vspace*{\baselineskip}
\end{table}

\subsection{Experiment Details}
\label{appendix:exp_detail}

\textbf{Experiment on the variables' influence (Fig.~\ref{fig:variables_analysis}).} We test the Macro F1 results of Falcon-7B-Instruct on SST-2, with: (1) 8 prompt templates with the same label space for the \textbf{Connector} experiment, (2) 8 prompt templates with only label space changed for the \textbf{Verbalizer} experiment, (3) 8 random seed for rearranging the input order for the \textbf{Order} experiment, and (4) 8 random seed for split the data into demonstration set and test set for the \textbf{Data Split} experiment.

\textbf{Implementation of calibration methods.} For methods that require additional ICL inference processing (Contextual Calibration, Domain Calibration, $k$-NN, Hidden Calibration), we provide 128 additional samples for the additional inference. For Domain Calibration, we fix the sample length to 64. For Batch Calibration, we fix the batch size to 128. For $k$-NN, we use the centroid form of k-NN to accelerate the inference.

\textbf{The unbiasedness of data sampling.} We evaluate sampling unbiasedness using two approaches: (1) We calculate the unbiasedness of the trisection by measuring the MAUVE~\citep{pillutla2021mauve} score averaged across tuples within the trisection set. The resulting score of $0.9802$ indicates a close distribution among these trisections. (2) We assess the unbiasedness of the sampled data from each subset by comparing it against the remaining data. Using the MAUVE score between the sampled and remaining data, we obtain a result of $0.9808$, further demonstrating minimal distributional deviation.

\begin{table}[t]
\setlength{\tabcolsep}{5pt}
\renewcommand{\arraystretch}{1.0}
\setlength{\abovecaptionskip}{2pt}
\centering
\caption{\small Alternative attributes under the meta-template for each dataset.}
\label{tab:alter}
\resizebox{1\columnwidth}{!}{
\begin{NiceTabular}{@{}cccccc}
\toprule \rowcolor{tablebg}
\textbf{Dataset} & \textbf{Index} & \textbf{Alternative Instruction} & \textbf{$x$ prefix} & \textbf{$y$ prefix} & \textbf{$y$ affix} \\ \midrule

\multirow{3}{*}{SST-2} & 0 & - & ``sentence: '' & ``sentiment: '' & ``$\backslash$n'' \\
                       & 1 & ``How would you describe the overall feeling of the movie based on this sentence? '' & ``text: '' & ``label: '' & `` '' \\
                       & 2 & ``Please classify the sentiment of the following sentence. '' & ``review: '' & ``Label: '' & ``$\backslash$t'' \\ \midrule
                       
\multirow{3}{*}{MR} & 0 & -  & ``review: '' & ``sentiment: '' & ``$\backslash$n''  \\
                       & 1 & ``How would you describe the overall feeling of the movie based on this sentence? '' & ``text: '' & ``label: '' & `` ''  \\
                       & 2 & ``Please classify the sentiment of the following sentence. '' & ``sentence: '' & ``Label: '' & ``$\backslash$t''  \\ \midrule
                       
\multirow{3}{*}{\thead{Financial\\Phrasebank}} & 0 & -  & ``sentence: '' & ``sentiment: '' & ``$\backslash$n'' \\
                       & 1 & ``What is the attitude towards the financial news in this sentence? '' & ``text: '' & ``label: '' & `` ''  \\
                       & 2 & ``What is the emotional response to the financial news in this sentence? '' & ``news: '' & ``Label: '' & ``$\backslash$t''  \\ \midrule
                       
\multirow{3}{*}{SST-5} & 0 & -  & ``sentence: '' & ``sentiment: '' & ``$\backslash$n''  \\
                       & 1 & ``How would you describe the overall feeling of the movie based on this sentence? ''  & ``text: '' & ``label: '' & `` ''  \\
                       & 2 & ``What mood does this sentence convey about the movie? '' & ``review: '' & ``Label: '' & ``$\backslash$t''  \\ \midrule
                       
\multirow{3}{*}{TREC} & 0 & -  & ``question: '' & ``target: '' & ``$\backslash$n''   \\
                       & 1 & ``What is the topic of the question? ''  & ``text: '' & ``label: '' & `` ''  \\
                       & 2 & ``What is the primary focus of this question? ''  & ``sentence: '' & ``Label: '' & ``$\backslash$t'' \\ \midrule
                       
\multirow{3}{*}{AGNews} & 0 & -  & ``news: '' & ``topic: '' & ``$\backslash$n''  \\
                       & 1 & ``What is the topic of the news? ''  & ``text: '' & ``label: '' & `` '' \\
                       & 2 & ``What is the news focused on? ''  & ``sentence: '' & ``Label: '' & ``$\backslash$t''\\ \midrule
                                              
\multirow{3}{*}{Subjective} & 0 & -  & ``review: '' & ``subjectiveness: '' & ``$\backslash$n''  \\
                       & 1 & ``Does this sentence reflect a personal opinion? ''  & ``text: '' & ``label: '' & `` '' \\
                       & 2 & ``Is this sentence expressing a personal opinion or stating a fact? ''  & ``sentence: '' & ``Label: '' & ``$\backslash$t''\\ \midrule
                                                                     
\multirow{3}{*}{\thead{Tweet\\Eval\\Emotion}} & 0 & -  & ``review: '' & ``subjectiveness: '' & ``$\backslash$n''  \\
                       & 1 & ``What feeling does this sentence convey? ''  & ``text: '' & ``label: '' & `` '' \\
                       & 2 & ``What emotion does this sentence express? ''  & ``sentence: '' & ``Label: '' & ``$\backslash$t''\\ \midrule
                                                                                            
\multirow{3}{*}{\thead{Tweet\\Eval\\Hate}} & 0 & -  & ``tweet: '' & ``hate speech: '' & ``$\backslash$n''  \\
                       & 1 & ``Does this sentence contain hate speech? ''  & ``text: '' & ``label: '' & `` '' \\
                       & 2 & ``Is this sentence an example of hate speech? ''  & ``sentence: '' & ``Label: '' & ``$\backslash$t''\\ \midrule
                                                                                                                   
\multirow{3}{*}{\thead{Hate\\Speech\\18}} & 0 & -  & ``tweet: '' & ``hate speech: '' & ``$\backslash$n''  \\
                       & 1 & ``Does this sentence contain hate speech? ''  & ``text: '' & ``label: '' & `` '' \\
                       & 2 & ``Is this sentence an example of hate speech? ''  & ``sentence: '' & ``Label: '' & ``$\backslash$t''\\
                       
\bottomrule
\end{NiceTabular}
}
\vspace*{-0.5\baselineskip}
\end{table}

\begin{table}[t]
\setlength{\tabcolsep}{5pt}
\renewcommand{\arraystretch}{1.0}
\setlength{\abovecaptionskip}{2pt}
\centering
\caption{\small Augmented results of Table~\ref{table:acc_improving}.}
\label{table:full_diag}
\resizebox{0.95\linewidth}{!}{
\begin{NiceTabular}{@{}cccccccccccccccc@{}}
\toprule \rowcolor{tablebg}
Models           & \textbf{GPT2}  & \textbf{Llama2} & \textbf{GPT2} & \textbf{GPT2} & \textbf{GPT2} & \textbf{Pythia} & \textbf{Pythia} & \textbf{Pythia} & \textbf{Pythia} & \textbf{Pythia} & \textbf{Pythia} & \textbf{Pythia} & \textbf{Pythia} & \textbf{OPT} & \textbf{OPT} \\  \rowcolor{tablebg}
Para. \# / B      & 0.137    & 6.74    & 0.38         & 0.812        & 1.61         & 0.014      & 0.07       & 0.16              & 0.41        & 1.08      & 1.52            & 2.91            & 6.9     & 2.7      & 1.3      \\ \midrule
Contextual Bias   & -0.8458  & -0.9717 & -0.9482      & -0.8281      & -0.9109      & -0.7927    & -0.7998    & -0.7722           & -0.7451     & -0.7519   & -0.7059         & -0.8645         & -0.7152                & -0.9050  & -0.9201  \\
Domain Bias       & -0.7737  & -0.9468 & -0.8516      & -0.7911      & -0.8749      & -0.7069    & -0.8072    & -0.7728           & -0.7497     & -0.7286   & -0.7027         & -0.8785         & -0.6922                & -0.8241  & -0.8834  \\
Empirical Bias      & 0.4302   & 0.1271  & 0.2123       & 0.3113       & 0.1214       & 0.4229     & 0.3322     & 0.2299            & 0.2629      & 0.2971    & 0.3395          & 0.2302          & 0.2905                 & 0.1512   & 0.1505   \\
Template Robust.     & 0.8702   & 0.8380  & 0.8046       & 0.8852       & 0.8052       & 0.8558     & 0.8823     & 0.8440            & 0.8878      & 0.9043    & 0.9084          & 0.9062          & 0.9134                 & 0.8873   & 0.8753   \\
Sample Robust.       & 0.8720   & 0.7594  & 0.8851       & 0.8851       & 0.8165       & 0.8558     & 0.7945     & 0.7428            & 0.7827      & 0.8266    & 0.8258          & 0.7924          & 0.8448                 & 0.8094   & 0.7350   \\
GLER             & 0.0127   & 0.0653  & 0.0317       & 0.0397       & 0.0403       & 0.0131     & 0.0244     & 0.0315            & 0.0448      & 0.0372    & 0.0426          & 0.0548          & 0.0413                 & 0.0471   & 0.0595   \\ \midrule \rowcolor{tablebg}
Models           & \textbf{OPT} & \textbf{GPT-J}  & \textbf{NEO} & \textbf{NEO} & \textbf{NEO} & \textbf{Falcon} & \textbf{Falcon*} & \textbf{Qwen2} & \textbf{Qwen} & \textbf{Qwen2}  & \textbf{Llama2} & \textbf{Llama2} & \textbf{Qwen2*} & \textbf{Llama3}  &          \\ \rowcolor{tablebg}
Para. \# / B     & 6.7      & 6.05    & 1.3          & 2.7          & 0.125        & 7.22       & 7.22       & 7.6               & 7.7         & 7.6       & 13              & 34              & 72                     & 8        &          \\ \midrule
Contextual Bias   & -0.9187  & -0.8878 & -0.9416      & -0.8719      & -0.8978      & -0.7992    & -0.9065    & -0.7429           & -0.7182     & -0.7747   & -0.9611         & -0.9743         & -0.7071                & -0.7771  &          \\
Domain Bias       & -0.9134  & -0.8269 & -0.7754      & -0.7140      & -0.7601      & -0.7679    & -0.9088    & -0.7557           & -0.6303     & -0.7602   & -0.9246         & -0.9222         & -0.6930                & -0.7730  &          \\
Empirical Bias      & 0.0848   & 0.1773  & 0.1489       & 0.3234       & 0.1371       & 0.2062     & 0.0513     & 0.1512            & 0.3439      & 0.2498    & 0.0921          & 0.0663          & 0.1143                 & 0.1401   &          \\
Template Robust.     & 0.8519   & 0.8648  & 0.8849       & 0.9051       & 0.8419       & 0.8746     & 0.8626     & 0.8741            & 0.9023      & 0.8973    & 0.8871          & 0.8243          & 0.8771                 & 0.8850   &          \\
Sample Robust.       & 0.7906   & 0.8019  & 0.7710       & 0.8253       & 0.6973       & 0.8795     & 0.8100     & 0.8763            & 0.9018      & 0.9278    & 0.8277          & 0.8010          & 0.9302                 & 0.8415   &          \\
GLER         & 0.0615   & 0.0239  & 0.0346       & 0.0331       & 0.0368       & 0.0406     & 0.0488     & 0.0725            & 0.0546      & 0.0630    & 0.0552          & 0.0689          & 0.0986                 & 0.0848   &          \\ \bottomrule
\end{NiceTabular}
}

\end{table}

\begin{table}[t]
\setlength{\tabcolsep}{5pt}
\renewcommand{\arraystretch}{1.0}
\setlength{\abovecaptionskip}{2pt}
\centering
\caption{\small Augmented results of Table~\ref{table:acc_improving}.}
\label{table:full_normal}
\resizebox{0.95\linewidth}{!}{
\begin{NiceTabular}{@{}ccccccccccccccccc@{}}
\toprule \rowcolor{tablebg}
 & \textbf{Model} & \textbf{GPT2}  & \textbf{Llama2} & \textbf{GPT2} & \textbf{GPT2} & \textbf{GPT2} & \textbf{Pythia} & \textbf{Pythia} & \textbf{Pythia} & \textbf{Pythia} & \textbf{Pythia} & \textbf{Pythia} & \textbf{Pythia} & \textbf{Pythia} & \textbf{OPT} & \textbf{OPT}  \\ \rowcolor{tablebg} 
\multirow{-2}{*}{\textbf{Method}} & {Para.\ \# / B }& 0.137    & 6.74    & 0.38         & 0.812        & 1.61         & 0.014      & 0.07       & 0.16              & 0.41        & 1.08      & 1.52            & 2.91            & 6.9                    & 2.7      & 1.3      \\ \midrule
\multirow{4}{*}{\textbf{\thead{Vanilla\\ICL}}} & \textbf{Accuracy}           & 0.4024   & 0.5732  & 0.4610       & 0.4583       & 0.5031       & 0.4045     & 0.3812     & 0.4490            & 0.4832      & 0.4507    & 0.4730          & 0.5065          & 0.5179                 & 0.5148   & 0.5021   \\
& \textbf{ATL}           & 0.3836   & 0.5360  & 0.4343       & 0.4478       & 0.4813       & 0.3923     & 0.3791     & 0.4313            & 0.4569      & 0.4412    & 0.4611          & 0.4865          & 0.5050                 & 0.4993   & 0.4790   \\
& \textbf{F1}            & 0.2863   & 0.5034  & 0.3503       & 0.3013       & 0.3411       & 0.2726     & 0.2747     & 0.3437            & 0.3493      & 0.3127    & 0.3416          & 0.3720          & 0.3426                 & 0.3647   & 0.3756   \\
& \textbf{ECE-1}          & 0.3353   & 0.1509  & 0.2267       & 0.3218       & 0.2411       & 0.2597     & 0.3337     & 0.2679            & 0.2650      & 0.3142    & 0.3082          & 0.2394          & 0.2866                 & 0.2546   & 0.2594   \\ \midrule

\multirow{4}{*}{\textbf{\thead{Noisy\\Channel}}} & \textbf{Accuracy}           & 0.4797   & 0.6135  & 0.4891       & 0.5304       & 0.5212       & 0.4285     & 0.4391     & 0.4499            & 0.4779      & 0.5240    & 0.5518          & 0.5669          & 0.5758                 & 0.5678   & 0.5449   \\
& \textbf{ATL}           & 0.3481   & 0.3484  & 0.3477       & 0.3484       & 0.3480       & 0.3473     & 0.3476     & 0.3475            & 0.3475      & 0.3482    & 0.3482          & 0.3480          & 0.3484                 & 0.3483   & 0.3481   \\
& \textbf{F1}            & 0.3983   & 0.5375  & 0.4096       & 0.4394       & 0.4345       & 0.3416     & 0.3559     & 0.3780            & 0.4130      & 0.4340    & 0.4610          & 0.4895          & 0.4866                 & 0.4699   & 0.4473   \\
& \textbf{ECE-1}          & 0.1281   & 0.2631  & 0.1383       & 0.1791       & 0.1705       & 0.0907     & 0.0955     & 0.1034            & 0.1385      & 0.1730    & 0.2010          & 0.2167          & 0.2252                 & 0.2171   & 0.1942   \\ \midrule

\multirow{4}{*}{\textbf{\thead{Contextual\\Calibration}}} &  \textbf{Accuracy}           & 0.3973   & 0.5200  & 0.3905       & 0.3814       & 0.4227       & 0.3586     & 0.3823     & 0.4573            & 0.4488      & 0.4335    & 0.4912          & 0.5056          & 0.4937                 & 0.4343   & 0.4410   \\
& \textbf{ATL}           & 0.3615   & 0.4675  & 0.3672       & 0.3603       & 0.3801       & 0.3399     & 0.3638     & 0.4242            & 0.3911      & 0.3880    & 0.4181          & 0.4309          & 0.4167                 & 0.4002   & 0.4021   \\
& \textbf{F1}            & 0.3421   & 0.5136  & 0.3485       & 0.3525       & 0.3956       & 0.2963     & 0.3459     & 0.3491            & 0.3856      & 0.3976    & 0.4125          & 0.4609          & 0.4445                 & 0.3798   & 0.3891   \\
& \textbf{ECE-1}          & 0.2366   & 0.2228  & 0.2321       & 0.2956       & 0.2607       & 0.1907     & 0.2685     & 0.2375            & 0.2274      & 0.1820    & 0.2189          & 0.1712          & 0.2189                 & 0.2340   & 0.2487   \\ \midrule

\multirow{4}{*}{\textbf{\thead{Domain\\Calibration}}} & \textbf{Accuracy}           & 0.4589   & 0.5977  & 0.4411       & 0.4520       & 0.4799       & 0.4257     & 0.4248     & 0.4726            & 0.4612      & 0.4817    & 0.4996          & 0.5457          & 0.5579                 & 0.5446   & 0.5022   \\
& \textbf{ATL}           & 0.4084   & 0.5087  & 0.3902       & 0.3948       & 0.4112       & 0.3915     & 0.4031     & 0.4423            & 0.3991      & 0.4103    & 0.4263          & 0.4532          & 0.4392                 & 0.4369   & 0.4248   \\
& \textbf{F1}            & 0.3904   & 0.5606  & 0.4059       & 0.4082       & 0.4189       & 0.3369     & 0.3661     & 0.3657            & 0.3951      & 0.4125    & 0.4211          & 0.4784          & 0.4667                 & 0.4738   & 0.4320   \\
& \textbf{ECE-1}          & 0.1277   & 0.1519  & 0.1594       & 0.1885       & 0.1918       & 0.1655     & 0.2021     & 0.2106            & 0.2172      & 0.1997    & 0.2022          & 0.1772          & 0.1992                 & 0.2265   & 0.2507   \\ \midrule

\multirow{4}{*}{\textbf{\thead{Batch\\Calibration}}} & \textbf{Accuracy}           & 0.4347   & 0.6057  & 0.4564       & 0.4748       & 0.5108       & 0.3912     & 0.4146     & 0.4351            & 0.4641      & 0.4642    & 0.4909          & 0.5345          & 0.5443                 & 0.5393   & 0.5034   \\
& \textbf{ATL}           & 0.3531   & 0.3954  & 0.3558       & 0.3575       & 0.3597       & 0.3461     & 0.3517     & 0.3534            & 0.3563      & 0.3550    & 0.3602          & 0.3674          & 0.3639                 & 0.3643   & 0.3626   \\
& \textbf{F1}            & 0.3896   & 0.5594  & 0.3968       & 0.4126       & 0.4468       & 0.3309     & 0.3708     & 0.3657            & 0.3903      & 0.3922    & 0.4066          & 0.4684          & 0.4469                 & 0.4616   & 0.4246   \\
& \textbf{ECE-1}          & 0.0678   & 0.2072  & 0.0928       & 0.0987       & 0.1427       & 0.0599     & 0.0677     & 0.0780            & 0.0855      & 0.0943    & 0.1049          & 0.1480          & 0.1753                 & 0.1644   & 0.1396   \\ \midrule
 
\multirow{4}{*}{\textbf{\thead{PPL-ICL}}} & \textbf{Accuracy}           & 0.4030   & 0.5834  & 0.4654       & 0.4609       & 0.5113       & 0.3973     & 0.3766     & 0.4547            & 0.4821      & 0.4521    & 0.4909          & 0.5045          & 0.5284                 & 0.5173   & 0.5004   \\
& \textbf{ATL}           & 0.3839   & 0.5412  & 0.4345       & 0.4457       & 0.4854       & 0.3890     & 0.3749     & 0.4357            & 0.4596      & 0.4455    & 0.4702          & 0.4884          & 0.5170                 & 0.4989   & 0.4736   \\
& \textbf{F1}            & 0.2898   & 0.5226  & 0.3513       & 0.3119       & 0.3581       & 0.2704     & 0.2704     & 0.3446            & 0.3447      & 0.3084    & 0.3627          & 0.3635          & 0.3551                 & 0.3727   & 0.3765   \\
& \textbf{ECE-1}          & 0.3198   & 0.1432  & 0.2146       & 0.3079       & 0.2247       & 0.2685     & 0.3354     & 0.2584            & 0.2721      & 0.3087    & 0.2898          & 0.2449          & 0.2853                 & 0.2412   & 0.2595   \\ \midrule

\multirow{4}{*}{\textbf{\thead{TopK}}} & \textbf{Accuracy}           & 0.4389   & 0.6660  & 0.5162       & 0.5730       & 0.6273       & 0.4391     & 0.5098     & 0.5381            & 0.5994      & 0.5777    & 0.6174          & 0.6430          & 0.6309                 & 0.6701   & 0.6883   \\
& \textbf{ATL}           & 0.4203   & 0.6053  & 0.4742       & 0.5511       & 0.5902       & 0.4170     & 0.4761     & 0.5054            & 0.5619      & 0.5459    & 0.5759          & 0.5969          & 0.5954                 & 0.6410   & 0.6475   \\
& \textbf{F1}            & 0.3358   & 0.6134  & 0.3963       & 0.4403       & 0.4869       & 0.3039     & 0.3975     & 0.4565            & 0.4777      & 0.4802    & 0.4957          & 0.5213          & 0.4876                 & 0.5715   & 0.5716   \\
& \textbf{ECE-1}          & 0.3191   & 0.1161  & 0.1990       & 0.2237       & 0.1914       & 0.2422     & 0.2613     & 0.2039            & 0.1968      & 0.2195    & 0.2026          & 0.1795          & 0.2239                 & 0.1715   & 0.1575   \\ \midrule

\multirow{4}{*}{\textbf{\thead{SA-ICL}}} & \textbf{Accuracy}           & 0.4385   & 0.6818  & -            & -            & 0.6328       & -          & -          & -                 & -           & -         & -               & -               & -                      & 0.6924   & 0.6945   \\
& \textbf{ATL}           & 0.4256   & 0.6337  & -            & -            & 0.6095       & -          & -          & -                 & -           & -         & -               & -               & -                      & 0.6690   & 0.6714   \\
& \textbf{F1}            & 0.3344   & 0.6131  & -            & -            & 0.4934       & -          & -          & -                 & -           & -         & -               & -               & -                      & 0.5808   & 0.5883   \\
& \textbf{ECE-1}          & 0.3707   & 0.1551  & -            & -            & 0.2525       & -          & -          & -                 & -           & -         & -               & -               & -                      & 0.2129   & 0.2230   \\ \midrule

\multirow{4}{*}{\textbf{\thead{$k$-NN}}} & \textbf{Accuracy}           & 0.3598   & 0.5642  & 0.4072       & 0.5609       & 0.5917       & 0.3782     & 0.3783     & 0.4096            & 0.5931      & 0.6035    & 0.6589          & 0.6998          & 0.7188                 & 0.5534   & 0.5192   \\
& \textbf{ATL}           & 0.3655   & 0.5567  & 0.4031       & 0.5519       & 0.5842       & 0.3714     & 0.3762     & 0.4075            & 0.5841      & 0.5908    & 0.6545          & 0.6895          & 0.7102                 & 0.5496   & 0.5155   \\
& \textbf{F1}            & 0.3138   & 0.5012  & 0.3278       & 0.4736       & 0.5110       & 0.3276     & 0.3189     & 0.3392            & 0.5121      & 0.5237    & 0.5702          & 0.6252          & 0.6436                 & 0.4791   & 0.4426   \\
& \textbf{ECE-1}         & 0.1633   & 0.2650  & 0.0847       & 0.3467       & 0.3445       & 0.1904     & 0.1185     & 0.0279            & 0.3374      & 0.3115    & 0.2849          & 0.2292          & 0.2211                 & 0.3812   & 0.4041   \\ \midrule

\multirow{4}{*}{\textbf{\thead{Hidden\\Calibration}}} & \textbf{Accuracy}           & 0.3842   & 0.6774  & 0.4087       & 0.6095       & 0.6339       & 0.4749     & 0.5121     & 0.4630            & 0.6132      & 0.6387    & 0.6709          & 0.7081          & 0.7179                 & 0.6319   & 0.5953   \\
& \textbf{ATL}           & 0.3782   & 0.6532  & 0.4066       & 0.4799       & 0.4971       & 0.4221     & 0.4524     & 0.4375            & 0.5581      & 0.5607    & 0.6100          & 0.6351          & 0.6608                 & 0.5527   & 0.5161   \\
& \textbf{F1}            & 0.3445   & 0.6101  & 0.3303       & 0.5119       & 0.5467       & 0.4279     & 0.4608     & 0.3931            & 0.5276      & 0.5576    & 0.5807          & 0.6305          & 0.6453                 & 0.5478   & 0.5113   \\
& \textbf{ECE-1}          & 0.4040   & 0.2084  & 0.4390       & 0.0924       & 0.1028       & 0.1194     & 0.1028     & 0.2905            & 0.1111      & 0.0618    & 0.0981          & 0.0647          & 0.0793                 & 0.0768   & 0.1042   \\ \midrule \rowcolor{tablebg} 
 &  \textbf{Model}  & \textbf{OPT} & \textbf{GPT-J}  & \textbf{NEO} & \textbf{NEO} & \textbf{NEO} & \textbf{Falcon} & \textbf{Falcon*} & \textbf{Qwen2} & \textbf{Qwen} & \textbf{Qwen2}  & \textbf{Llama2} & \textbf{Llama2} & \textbf{Qwen2*} & \textbf{Llama3} &  \\ \rowcolor{tablebg} \multirow{-2}{*}{\textbf{Method}}
 & {Para.\ \# / B}  & 6.7      & 6.05    & 1.3          & 2.7          & 0.125        & 7.22       & 7.22       & 7.6               & 7.7         & 7.6       & 13              & 34              & 72                     & 8        &          \\ \midrule

\multirow{4}{*}{\textbf{\thead{Vanilla\\ICL}}} & \textbf{Accuracy}           & 0.5992   & 0.5341  & 0.4840       & 0.4435       & 0.4477       & 0.5691     & 0.6204     & 0.6687            & 0.5796      & 0.6119    & 0.6347          & 0.6683          & 0.7236                 & 0.6442   &          \\
& \textbf{ATL}           & 0.5355   & 0.4994  & 0.4654       & 0.4415       & 0.4294       & 0.5219     & 0.5522     & 0.6254            & 0.5596      & 0.5590    & 0.5722          & 0.6099          & 0.7000                 & 0.5872   &          \\
& \textbf{F1}            & 0.4756   & 0.4039  & 0.3457       & 0.3245       & 0.3487       & 0.4382     & 0.5240     & 0.5570            & 0.4537      & 0.4689    & 0.5263          & 0.5950          & 0.6360                 & 0.5309   &          \\
& \textbf{ECE-1}         & 0.1725   & 0.2467  & 0.2462       & 0.3425       & 0.2889       & 0.2025     & 0.1076     & 0.1461            & 0.2897      & 0.2245    & 0.1661          & 0.1237          & 0.1585                 & 0.1650   &          \\ \midrule

\multirow{4}{*}{\textbf{\thead{Noisy\\Channel}}} & \textbf{Accuracy}           & 0.5731   & 0.5897  & 0.5449       & 0.5678       & 0.4586       & 0.6069     & 0.6111     & 0.6376            & 0.5709      & 0.6371    & 0.6496          & 0.6111          & 0.6872                 & 0.6647   &          \\
& \textbf{ATL}           & 0.3484   & 0.3485  & 0.3481       & 0.3483       & 0.3477       & 0.3483     & 0.3485     & 0.3502            & 0.3487      & 0.3488    & 0.3490          & 0.3496          & 0.3512                 & 0.3494   &          \\
& \textbf{F1}            & 0.4844   & 0.5149  & 0.4473       & 0.4699       & 0.3832       & 0.5317     & 0.5201     & 0.5589            & 0.4792      & 0.5574    & 0.5718          & 0.5390          & 0.6175                 & 0.5867   &          \\
& \textbf{ECE-1}          & 0.2234   & 0.2391  & 0.1942       & 0.2171       & 0.1129       & 0.2569     & 0.2607     & 0.2853            & 0.2205      & 0.2867    & 0.2988          & 0.2592          & 0.3341                 & 0.3136   &          \\ \midrule

\multirow{4}{*}{\textbf{\thead{Contextual\\Calibration}}} & \textbf{Accuracy}           & 0.5304   & 0.5487  & 0.4701       & 0.4343       & 0.3837       & 0.5019     & 0.5485     & 0.6404            & 0.5588      & 0.5923    & 0.5707          & 0.6280          & 0.7143                 & 0.5544   &          \\
& \textbf{ATL}           & 0.4519   & 0.4634  & 0.4186       & 0.4002       & 0.3579       & 0.4398     & 0.4748     & 0.5533            & 0.4939      & 0.5055    & 0.4880          & 0.5458          & 0.6039                 & 0.4825   &          \\
& \textbf{F1}            & 0.4812   & 0.4657  & 0.3808       & 0.3798       & 0.3555       & 0.4906     & 0.5395     & 0.6089            & 0.5277      & 0.5527    & 0.5330          & 0.5808          & 0.6730                 & 0.5214   &          \\
& \textbf{ECE-1}          & 0.2087   & 0.1761  & 0.1935       & 0.2340       & 0.3329       & 0.2067     & 0.2258     & 0.2122            & 0.1829      & 0.2029    & 0.2013          & 0.1418          & 0.1854                 & 0.2692   &          \\ \midrule

\multirow{4}{*}{\textbf{\thead{Domain\\Calibration}}} & \textbf{Accuracy}           & 0.5525   & 0.5528  & 0.4635       & 0.4783       & 0.4450       & 0.5457     & 0.5725     & 0.6558            & 0.6171      & 0.6190    & 0.6238          & 0.6657          & 0.7148                 & 0.6240   &          \\
& \textbf{ATL}           & 0.4585   & 0.4588  & 0.4199       & 0.4260       & 0.4135       & 0.4698     & 0.4877     & 0.5531            & 0.5177      & 0.5161    & 0.5175          & 0.5692          & 0.6048                 & 0.5149   &          \\
& \textbf{F1}            & 0.4945   & 0.4921  & 0.3912       & 0.4375       & 0.3888       & 0.5327     & 0.5204     & 0.6283            & 0.5543      & 0.5788    & 0.5877          & 0.6219          & 0.6692                 & 0.5595   &          \\
& \textbf{ECE-1}          & 0.2117   & 0.1640  & 0.2107       & 0.1963       & 0.2640       & 0.1919     & 0.2044     & 0.1599            & 0.1891      & 0.1524    & 0.1615          & 0.1461          & 0.1810                 & 0.1635   &          \\ \midrule

\multirow{4}{*}{\textbf{\thead{Batch\\Calibration}}} & \textbf{Accuracy}           & 0.5861   & 0.5554  & 0.4734       & 0.5011       & 0.4340       & 0.6084     & 0.6229     & 0.6952            & 0.6418      & 0.6590    & 0.6280          & 0.6663          & 0.7345                 & 0.6589   &          \\
& \textbf{ATL}           & 0.3802   & 0.3732  & 0.3586       & 0.3607       & 0.3555       & 0.3831     & 0.3902     & 0.4132            & 0.3978      & 0.3946    & 0.4014          & 0.4182          & 0.4424                 & 0.3959   &          \\
& \textbf{F1}            & 0.5133   & 0.4954  & 0.4088       & 0.4533       & 0.3790       & 0.5569     & 0.5668     & 0.6288            & 0.5665      & 0.5946    & 0.5726          & 0.6142          & 0.6687                 & 0.5733   &          \\
& \textbf{ECE-1}          & 0.1829   & 0.1680  & 0.0890       & 0.1413       & 0.0719       & 0.2182     & 0.2298     & 0.2596            & 0.2306      & 0.2527    & 0.2168          & 0.2081          & 0.2608                 & 0.2419   &          \\ \midrule

\multirow{4}{*}{\textbf{\thead{PPL-ICL}}} & \textbf{Accuracy}           & 0.6036   & 0.5438  & 0.4738       & 0.4241       & 0.4538       & 0.5642     & 0.6331     & 0.6973            & 0.5865      & 0.6559    & 0.6240          & 0.6921          & 0.7581                 & 0.6625   &          \\
& \textbf{ATL}           & 0.5403   & 0.4932  & 0.4541       & 0.4209       & 0.4337       & 0.5160     & 0.5665     & 0.6603            & 0.5614      & 0.5851    & 0.5666          & 0.6326          & 0.7388                 & 0.5944   &          \\
& \textbf{F1}            & 0.4811   & 0.4279  & 0.3358       & 0.3128       & 0.3431       & 0.4314     & 0.5484     & 0.6086            & 0.4632      & 0.5465    & 0.5314          & 0.6283          & 0.6705                 & 0.5484   &          \\
& \textbf{ECE-1}         & 0.1701   & 0.2342  & 0.2535       & 0.3560       & 0.2882       & 0.2016     & 0.0952     & 0.1398            & 0.2789      & 0.1928    & 0.1745          & 0.1161          & 0.1485                 & 0.1721   &          \\ \midrule

\multirow{4}{*}{\textbf{\thead{TopK}}} & \textbf{Accuracy}           & 0.7293   & 0.6311  & 0.6018       & 0.5391       & 0.5656       & 0.6164     & 0.6988     & 0.6820            & 0.6283      & 0.6703    & 0.7375          & 0.7252          & 0.7006                 & 0.7625   &          \\
& \textbf{ATL}           & 0.6681   & 0.6044  & 0.5714       & 0.5325       & 0.5451       & 0.5662     & 0.6243     & 0.6466            & 0.6152      & 0.6153    & 0.6582          & 0.6661          & 0.6843                 & 0.6861   &          \\
& \textbf{F1}            & 0.6492   & 0.5352  & 0.4919       & 0.4257       & 0.4726       & 0.4855     & 0.6187     & 0.5910            & 0.5283      & 0.5648    & 0.6556          & 0.6595          & 0.6328                 & 0.6708   &          \\
& \textbf{ECE-1}          & 0.1301   & 0.1923  & 0.1644       & 0.2739       & 0.2085       & 0.1844     & 0.0796     & 0.1519            & 0.2520      & 0.2025    & 0.1299          & 0.1005          & 0.1787                 & 0.1476   &          \\ \midrule

\multirow{4}{*}{\textbf{\thead{SA-ICL}}} & \textbf{Accuracy}           & 0.7566   & -       & -            & -            & -            & 0.6146     & 0.7152     & 0.6934            & -           & 0.6754    & 0.7537          & 0.7434          & 0.7078                 & 0.7713   &          \\
& \textbf{ATL}           & 0.7125   & -       & -            & -            & -            & 0.5771     & 0.6588     & 0.6690            & -           & 0.6331    & 0.6960          & 0.7135          & 0.7038                 & 0.7091   &          \\
& \textbf{F1}            & 0.6804   & -       & -            & -            & -            & 0.4876     & 0.6345     & 0.5991            & -           & 0.5671    & 0.6696          & 0.6746          & 0.6416                 & 0.6757   &          \\
& \textbf{ECE-1}          & 0.1618   & -       & -            & -            & -            & 0.2412     & 0.1259     & 0.2066            & -           & 0.2277    & 0.1633          & 0.1699          & 0.2360                 & 0.1734   &          \\ \midrule

\multirow{4}{*}{\textbf{\thead{$k$-NN}}} & \textbf{Accuracy}           & 0.6044   & 0.6544  & 0.4896       & 0.5137       & 0.4490       & 0.6531     & 0.7029     & 0.7352            & 0.7162      & 0.6710    & 0.6732          & 0.6937          & 0.7090                 & 0.6477   &          \\
& \textbf{ATL}           & 0.6020   & 0.6493  & 0.4878       & 0.5090       & 0.4473       & 0.6425     & 0.6977     & 0.7244            & 0.7163      & 0.6581    & 0.6621          & 0.6866          & 0.7053                 & 0.6454   &          \\
& \textbf{F1}            & 0.5271   & 0.6012  & 0.4386       & 0.4544       & 0.3907       & 0.5754     & 0.6382     & 0.6728            & 0.6439      & 0.6054    & 0.6010          & 0.6289          & 0.6400                 & 0.5624   &          \\
& \textbf{ECE-1}          & 0.3396   & 0.2864  & 0.4229       & 0.4048       & 0.4494       & 0.2645     & 0.2554     & 0.1648            & 0.2230      & 0.1854    & 0.2456          & 0.2532          & 0.1624                 & 0.2350   &          \\ \midrule

\multirow{4}{*}{\textbf{\thead{Hidden\\Calibration}}} & \textbf{Accuracy}           & 0.6929   & 0.7066  & 0.5547       & 0.6057       & 0.4802       & 0.6483     & 0.6955     & 0.8053            & 0.7667      & 0.7594    & 0.7012          & 0.6891          & 0.8070                 & 0.7356   &          \\
& \textbf{ATL}           & 0.6256   & 0.6364  & 0.5265       & 0.5677       & 0.4554       & 0.5521     & 0.6155     & 0.7914            & 0.7609      & 0.7441    & 0.6511          & 0.6464          & 0.7991                 & 0.7148   &          \\
& \textbf{F1}            & 0.6093   & 0.6532  & 0.4955       & 0.5412       & 0.4263       & 0.5739     & 0.6303     & 0.7436            & 0.6906      & 0.6908    & 0.6288          & 0.6230          & 0.7438                 & 0.6540   &          \\
& \textbf{ECE-1}          & 0.0823   & 0.0967  & 0.1792       & 0.1661       & 0.2136       & 0.1139     & 0.1055     & 0.1481            & 0.1780      & 0.1785    & 0.1400          & 0.1426          & 0.1565 & 0.1845 \\ \bottomrule
\end{NiceTabular}
}

\end{table}

\section{Experiment Result Details}
\label{appendix:more_res}

The augmented results of \MN~in Table~\ref{table:acc_improving} for all models and metrics are shown in Table~\ref{table:full_normal}.

The augmented results of \MD~in Table~\ref{table:diag} for all models and metrics are shown in Table~\ref{table:full_diag}.

\end{document}